\documentclass[runningheads]{llncs}

\usepackage{eccv}

\usepackage{eccvabbrv}

\usepackage{graphicx}
\usepackage{booktabs}

\usepackage{wrapfig}
\usepackage{amsmath,graphicx}
\usepackage{setspace}
\usepackage{booktabs}
\usepackage{bbding}
\usepackage{epstopdf}
\usepackage{multirow}
\usepackage{color}
\usepackage{url}
\usepackage{caption}
\usepackage{algorithmic}
\usepackage{algorithm}
\usepackage{bbding}

\usepackage[accsupp]{axessibility}  

\usepackage{hyperref}

\usepackage{orcidlink}

\newcommand{\blue}[1]{{\color{blue}#1}}

\makeatletter
\newcommand{\correspondingauthor}{\textsuperscript{\Envelope}}
\newcommand{\coauthor}{\textsuperscript{$\star$}}
\makeatother

\begin{document}
	
\title{Fake It till You Make It: Curricular Dynamic Forgery Augmentations towards General Deepfake Detection} 

\titlerunning{Curricular Dynamic Forgery Augmentations for Deepfake Detection}

\author{Yuzhen Lin \thanks{Equal contributions. ~~~ \Envelope~Corresponding author.} \inst{1}\orcidlink{0000-0001-7788-2054} \and
	Wentang Song \coauthor \inst{1}\orcidlink{0000-0003-3750-9516} \and
	Bin Li \correspondingauthor \inst{1}\orcidlink{0000-0002-2613-5451} \and
	Yuezun Li\inst{2}\orcidlink{0000-0001-9299-1945} \and
	Jiangqun Ni\inst{3}\orcidlink{0000-0002-7520-9031} \and
	Han Chen\inst{1}\orcidlink{0000-0002-9439-9133} \and
	Qiushi Li\inst{1}\orcidlink{0000-0002-4976-3346}
}

\authorrunning{Lin et al.}

\institute{Guangdong Provincial Key Laboratory of Intelligent Information Processing,  
	Shenzhen Key Laboratory of Media Security, 
	SZU-AFS Joint Innovation Center for AI Technology, 
	Shenzhen University, Shenzhen, China \\
	\email{\{linyuzhen2020, 2018132120, 2016130205, 1800271017\}@email.szu.edu.cn; \Envelope~libin@szu.edu.cn;}
	\and
	College of Computer Science and Technology, Ocean University of China, Qingdao, China; 
	\email{liyuezun@ouc.edu.cn}
	\and
	School of Cyber Science and Technology, Sun Yat-Sen University, and Department of New Networks, Peng Cheng Laboratory, Shenzhen, China \\
	\email{issjqni@mail.sysu.edu.cn}
}

\maketitle


\begin{abstract}
	Previous studies in deepfake detection have shown promising results when testing face forgeries from the same dataset as the training. 
	However, the problem remains challenging when one tries to generalize the detector to forgeries from unseen datasets and created by unseen methods. 
	In this work, we present a novel general deepfake detection method, called \textbf{C}urricular \textbf{D}ynamic \textbf{F}orgery \textbf{A}ugmentation (CDFA), which jointly trains a deepfake detector with a forgery augmentation policy network. 
	Unlike the previous works, we propose to progressively apply forgery augmentations following a monotonic curriculum during the training. 
	We further propose a dynamic forgery searching strategy to select one suitable forgery augmentation operation for each image varying between training stages, producing a forgery augmentation policy optimized for better generalization.  
	In addition, we propose a novel forgery augmentation named self-shifted blending image to simply imitate the temporal inconsistency of deepfake generation.
	Comprehensive experiments show that CDFA can significantly improve both cross-datasets and cross-manipulations performances of various naive deepfake detectors in a plug-and-play way, and make them attain superior performances over the existing methods in several benchmark datasets. 
\end{abstract}

\section{Introduction}
\label{sec:intro}

Deepfake techniques \cite{FSGANSubjectAgnostic2019nirkin,AdvancingHighFidelity2020li,InformationBottleneckDisentanglement2021gao,StyleSwapStyleBasedGenerator2022xu,DesigningOneUnified2022xu,FSGANv2ImprovedSubject2023nirkin} refer to a series of deep learning-based facial forgery techniques that can swap or reenact the face of one person in a video to another. It poses a significant threat given their potential by spreading false information and even political manipulation. 
To reduce these risks, detecting deepfakes has become a crucial research topic in recent years.

Early works \cite{MesoNetCompactFacial2018afchar,FaceForensicsLearningDetect2019rossler} treat deepfake detection as a binary classification problem and directly use deep neural networks \cite{XceptionDeepLearning2017chollet,EfficientNetRethinkingModel2019tan} to distinguish fake faces (named naive deepfake detectors \cite{DeepfakeBenchComprehensiveBenchmark2023yan}). In order to improve the detection performance, some works \cite{ThinkingFrequencyFace2020qian,MultiAttentionalDeepfakeDetection2021zhao,SpatialPhaseShallowLearning2021liu,GeneralizingFaceForgery2021luo,ExposingFaceForgery2022chen} introduce auxiliary modalities (e.g., frequency) or supervision (e.g., forgery masks) information for learning subtle forgery artifacts. These methods achieve promising performance in a closed-domain scenario, where the training and testing data are sampled from the same distribution. However, in practice the testing forgeries are usually from unseen datasets and synthesized by unseen methods. Discrepancies between training and testing data lead to inferior performance of detectors, which poses challenges to deepfake detectors for practical usage.

Recall that a forgery can be easily synthesized by blending two different images. Motivated by this, a powerful solution to improve the generalization capabilities of deepfake detectors is introducing the forgery augmentation technology \cite{FaceXRayMore2020li,DetectingDeepfakesSelfBlended2022shiohara} that blends two real faces from training data to get new face forgeries.
The augmented sample (labeled as fake) is so-called pseudo fake (\textit{p-fake}) sample \cite{DetectingDeepfakeCreating2023guan} to distinguish them from the original fake  (\textit{o-fake}) sample of the training data.
Forgery augmentation strategies are also at the core of many state-of-the-art (SOTA) detection models \cite{LearningSelfConsistencyDeepfake2021zhao,ProtectingCelebritiesDeepFake2022dong,OSTImprovingGeneralization2022chena,AUNetLearningRelations2023bai,SeeABLESoftDiscrepancies2023larue,LAANetLocalizedArtifact2024nguyen}.
One shared intuition among such methods is that they utilize forgery augmentations to imitate the deepfake generation pipeline to encourages detection models to learn generic representative features.

Despite the success of such forgery augmentation-based methods, most of them exploit p-fake samples for training models in only two ways: 1) utilizing solely p-fake samples without the incorporation of o-fake samples, or 2) creating some p-fake samples and then mixing them into o-fake samples. 
In other words, the number of p-fake samples and the policy of forgery augmentation are fixed when training the deepfake detector. 
It may lead to inefficient training for the following reasons:
First, applying forgery augmentation does not always bring improvement over the whole process of training. 
For instance, we observed that a detection model tends to learn faster during earlier training stages without using forgery augmentation. We hypothesize that models at the early stage of training still lacks the capability to recognize the original forgeries, so excessively introduced p-fake samples at such stages are not conducive to the convergence of the models.
Secondly, using only a single type of forgery augmentation scheme to generate p-fake samples during the training is not optimal for the model. 
Intuitively, the detection model can learn more clues from p-fake samples synthesized by diverse kinds of forgery augmentation operations. 
Moreover, the optimal forgery augmentation scheme should be different for every sample on variation of training stages.

\begin{figure}[t]
	\centering
	\includegraphics[width=0.9\linewidth]{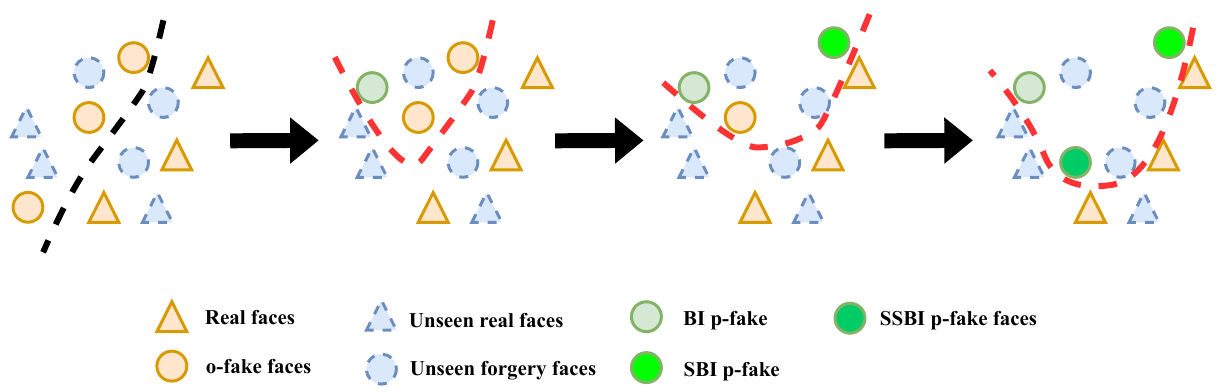}
	\caption{The proposed CDFA adjusts the composition of fake samples during the training by: 1) gradually increasing the proportion of p-fake samples, and 2) applying a dynamic forgery augmentation policy to generate p-fake samples.}
	\label{fig:motivation}
\end{figure}

Motivated by aforementioned concern, in this work, we propose a novel Curricular Dynamic Forgery Augmentation (CDFA) strategy. CDFA is a simple yet efficient method to improve the generalization for deepfake detectors by adjusting the composition of fake samples at different training stages (see Figure \ref{fig:motivation}). 
As for the number of p-fake samples, we design a \textit{Monotonic Curriculum (MC)} strategy that progressively introduces more p-fake samples while reducing the o-fake samples as training proceeds. 
Although the monotonic curriculum gradually increases the p-fake samples as the model improves, it does not determine which forgery augmentation operation applied to each sample can bring the most improvement to the model training. Motivated by the automatic augmentation paradigm \cite{AutoAugmentLearningAugmentation2019cubuk,SurveyAutomatedData2023cheung}, we propose a \textit{ Dynamic Forgery Search (DFS)} strategy which considers the evaluation of the current model on the validation set as an expert to guide the optimization of which forgery augmentation operation is applied to each sample in different training stages.
Furthermore, considering the current forgery augmentations \cite{FaceXRayMore2020li,DetectingDeepfakesSelfBlended2022shiohara} can not imitate the temporal inconsistency of deepfake generation, we propose a novel forgery augmentation named \textit{Self-shifted Blending Image (SSBI)}. It can simply introduce the temporal artifacts by blending the faces of two different frames from the same video. 
Comprehensive experimental results show that our method can significantly improve the generalization performances of various naive deepfake detectors in a plug-and-play manner, make them achieve superior performances over several SOTA competitors in multiple cross-datasets and cross-manipulations benchmarks. 

Briefly, the main contributions of this work can be summarized as follows:
\begin{itemize}
	\item To the best of our knowledge, it is the first work to investigate the p-fake sample scheme, including its proportion and generation method, during the training of deepfake detector.
	\item We propose a monotonic curriculum strategy that gradually introduces the proportion of p-fake samples along with the training process. 
	Through such easy-to-hard data strategy, we can improve the generalization performance while accelerating convergence of the deepfake detector.
	\item We propose a dynamic forgery search strategy that trains a policy network on the fly with the training of deepfake detector, which aims to search a optimal forgery augmentation policy based on evolving data and model states in different training stages.
	\item We futher propose a novel forgery augmentation method, named Self-shifted Blending Image (SSBI), to compensates the deficiency of prior works in simulating temporal artifacts.
\end{itemize}

\section{Related Works}
\label{sec:related}

\noindent \textbf{Deepfake detection.}
The past five years have witnessed a wide variety of methods proposed for defending against the malicious usage of deepfakes. 
Early works focus on hand-crafted features such as eyes-blinking \cite{IctuOculiExposing2018li}, inconsistent head poses \cite{ExposingDeepFakes2019yang} and visual artifacts \cite{ExploitingVisualArtifacts2019matern,ExposingDeepFakeVideos2019li}. By formulating the detecting as a vanilla binary classification problem (i.e. pristine or forgery), current end-to-end trained detectors \cite{MesoNetCompactFacial2018afchar,XceptionDeepLearning2017chollet,EfficientNetRethinkingModel2019tan} to directly distinguish deepfake content from authentic data. To this end, several works \cite{ThinkingFrequencyFace2020qian,SpatialPhaseShallowLearning2021liu,FrequencyAwareDiscriminativeFeature2021li} utilize frequency information to improve the performance of detectors. Moreover, there are some works aiming to localize the forged regions and make a decision based on the predicted regions \cite{ExposingFaceForgery2022chen}. Due to the development of deep generative models, the forged faces become more realistic and the manipulation methods are of more diversity. Some works propose to find clues on inconsistency of facial identity \cite{DeepFakeDetectionBased2022nirkin,ProtectingCelebritiesDeepFake2022dong,UnmaskingDeepfakesMasked2023das,ImplicitIdentityLeakage2023dong}.
Several works show introduce common data augmentations (e.g, blurring and jpeg compression) \cite{CNNGeneratedImagesAre2020wang,TrainingStrategiesData2020bondi,DeepfakeBenchComprehensiveBenchmark2023yan} can help improve the detection performance. Furthermore, \cite{ImprovingCrossdatasetGeneralization2022nadimpalli} proposes to use RL agent to search the policy of common data augmentations (e.g., Brightness and Contrast). 
However, the improvement in generalization performance of the commonly data augmentation is limited.

\noindent \textbf{Deepfake detection through forgery augmentation.}
One of the most effective approaches to improve generalization performance is to introduce forgery augmentation techniques to first synthesize forged images (i.e., pseudo deepfakes) and then train a deepfake detector model. As the pioneering works, BI \cite{FaceXRayMore2020li} are introduced to generate blended faces which reproduce blending artifacts from pairs of two pristine images with similar facial landmarks. Following that, 
SBI \cite{DetectingDeepfakesSelfBlended2022shiohara} selects two views of the same face image as the target face and the source face. SLADD \cite{SelfSupervisedLearningAdversarial2022chen} employs an adversarial training strategy to find the most difficult BI configuration and trained a classifier to predict the forgeries. Recent works \cite{LearningSelfConsistencyDeepfake2021zhao,ProtectingCelebritiesDeepFake2022dong,OSTImprovingGeneralization2022chena,AUNetLearningRelations2023bai,SeeABLESoftDiscrepancies2023larue,LAANetLocalizedArtifact2024nguyen} also conduct such forgery augmentation paradigm as a core part of improving generalization performances.

\section{Methodology}
\label{sec:method}

\begin{figure}[t]
	\centering
	\includegraphics[width=\linewidth]{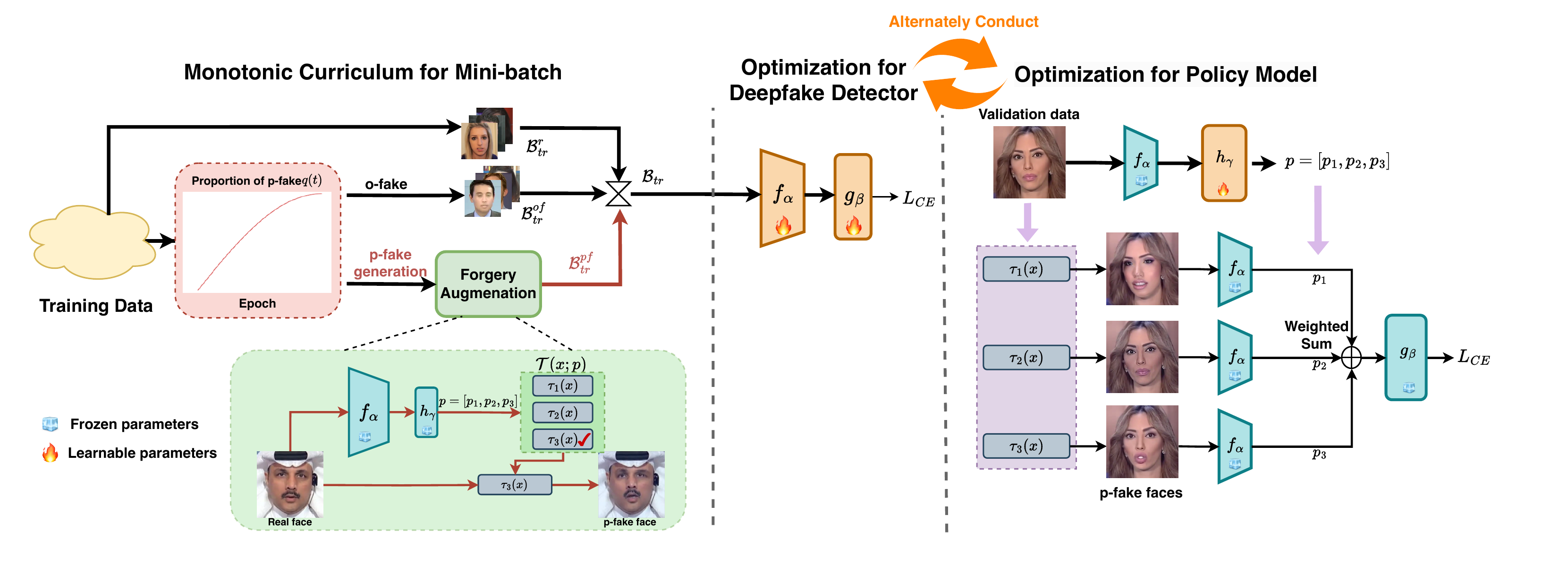}
	\caption{Overview of the proposed CDFA. }
	\label{fig:overall}
\end{figure}

In this section, we describe the proposed CDFA method in detail. The pipeline of CDFA is shown in Figure \ref{fig:overall}. 
First, we propose a monotonic curriculum strategy, designed to gradually increase the proportion of p-fake samples and decrease the proportion of of o-fake samples in each mini-batch as the training proceeds.
As for p-fake generation, we propose a dynamic forgery search strategy that optimizes a lightweight policy network to determine the preferred forgery augmentation operation for producing p-fake samples in different training stages.

\begin{wrapfigure}{r}{0.35\textwidth} 
	\centering
	\includegraphics[width=\linewidth]{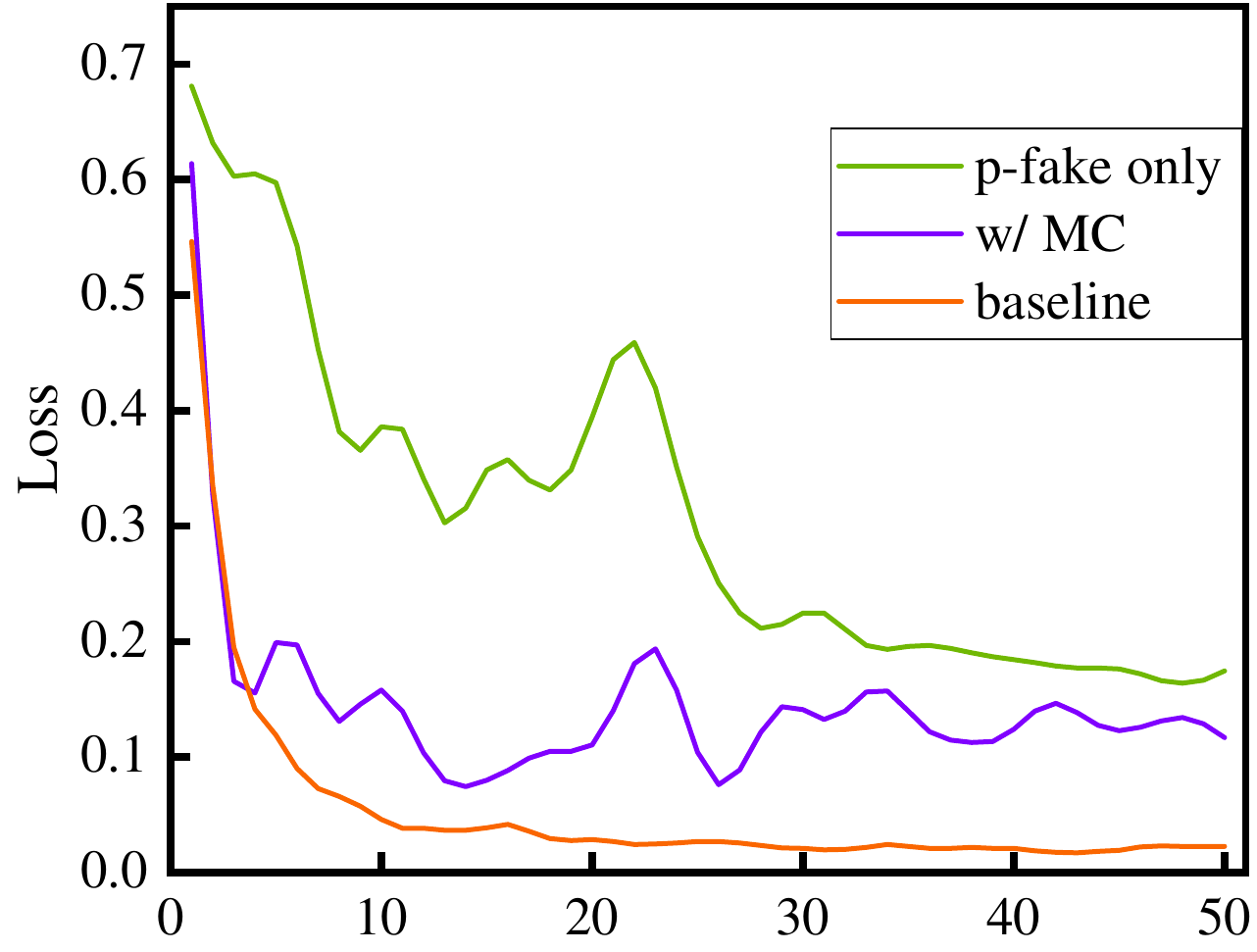}
	\caption{Validation loss on FF++ during the training.}
	\label{fig:loss_conv}
\end{wrapfigure}

\subsection{Monotonic Curriculum}
Previous works \cite{FaceXRayMore2020li,DetectingDeepfakesSelfBlended2022shiohara} of forgery augmentations are used to simply mix p-fake samples into o-fake samples or use them directly (without o-fake samples) and then conduct the model training. 
Herein, we conducted a simple study by training the model by using o-fake (i.e., baseline) or p-fake samples alone. 
Figure \ref{fig:loss_conv} shows that the model converges much slower when only trained with p-fake samples (generated by SBI\cite{DetectingDeepfakesSelfBlended2022shiohara}).
This suggests that the model can not even recognize the original forgery artifacts at the very early stage of the training.
In other words, for the deepfake detection task, the o-fake samples can be considered as easy samples, while the p-fake is more difficult.
Consequently, introducing a large number of p-fake samples at the initial stage appears to be not optimal for achieving efficient model convergence. 

Inspired by curriculum learning paradigm \cite{CurriculumLearning2009bengio,SurveyCurriculumLearning2022wang,CurriculumLearningSurvey2022soviany,WhenLearnWhat2023hou,GenericDeepfakeDetection2024song}, we propose a easy-to-hard data strategy that adjust the proportion of p-fake samples along with the training process.
We introduce the curriculum schedule $q(t)$ about the proportion of p-fake samples as follows:
\begin{equation}\label{eq:mc}
	q(t) = \sin(t/\epsilon)
\end{equation}
where $t$ is the current training epoch number and $\epsilon$ is a manually adjustable hyper-parameter. We set $\epsilon=2T/\pi$ to make $q(t)$ increases monotonically in $[0, 1]$, where $T$ is the total number of the training epoch. 

Let $\mathcal{D}^{r}_{tr}$ and $\mathcal{D}^{f}_{tr}$ be the real and fake part of the training data $\mathcal{D}_{tr}$, respectively.
In constructing a training mini-batch $\mathcal{B}_{tr}$ with batch size $b$, we first sample $b/2$ images from $\mathcal{D}^{r}_{tr}$ as the real part, denoted as $\mathcal{B}^{r}_{tr}$. For the fake part of $\mathcal{B}_{tr}$, we compute number of o-fake and p-fake samples (denoted as $n_{pf}$ and $n_{of}$ respectively) by:
\begin{equation}\label{eq:mcn}
	n_{pf}=q(t) \times b/2, n_{of}=(1-q(t)) \times b/2
\end{equation}
For o-fake samples,  $\mathcal{B}^{of}_{tr}$, we sample $n_{of}$ images from $\mathcal{D}^{f}_{tr}$. For p-fake samples $\mathcal{B}^{pf}_{tr}$, we random select $n_{pf}$ images from $\mathcal{D}^{r}_{tr}$ and conduct forgery augmentations to generate them. Thus, $\mathcal{B}_{tr}$ is obtained by:
\begin{equation}\label{eq:batch}
	\mathcal{B}_{tr}=\mathcal{B}^{r}_{tr} \cup \mathcal{B}^{of}_{tr} \cup \mathcal{B}^{pf}_{tr}, |\mathcal{B}_{tr}|=b
\end{equation}
Utilizing this strategy, the model is primarily trained on the o-fake samples at the early stages, which facilitates rapid convergence by learning the obvious forgery traces in o-fake samples. As training proceeds, the model fully learns the original forgery artifacts and its training can benefit more from the augmented p-fake samples.
To verify this, we observe that the convergence efficiency of the model (see in Figure \ref{fig:loss_conv}) becomes higher after introducing the MC strategy.

\begin{figure}[t]
	\centering
	\includegraphics[width=0.6\linewidth]{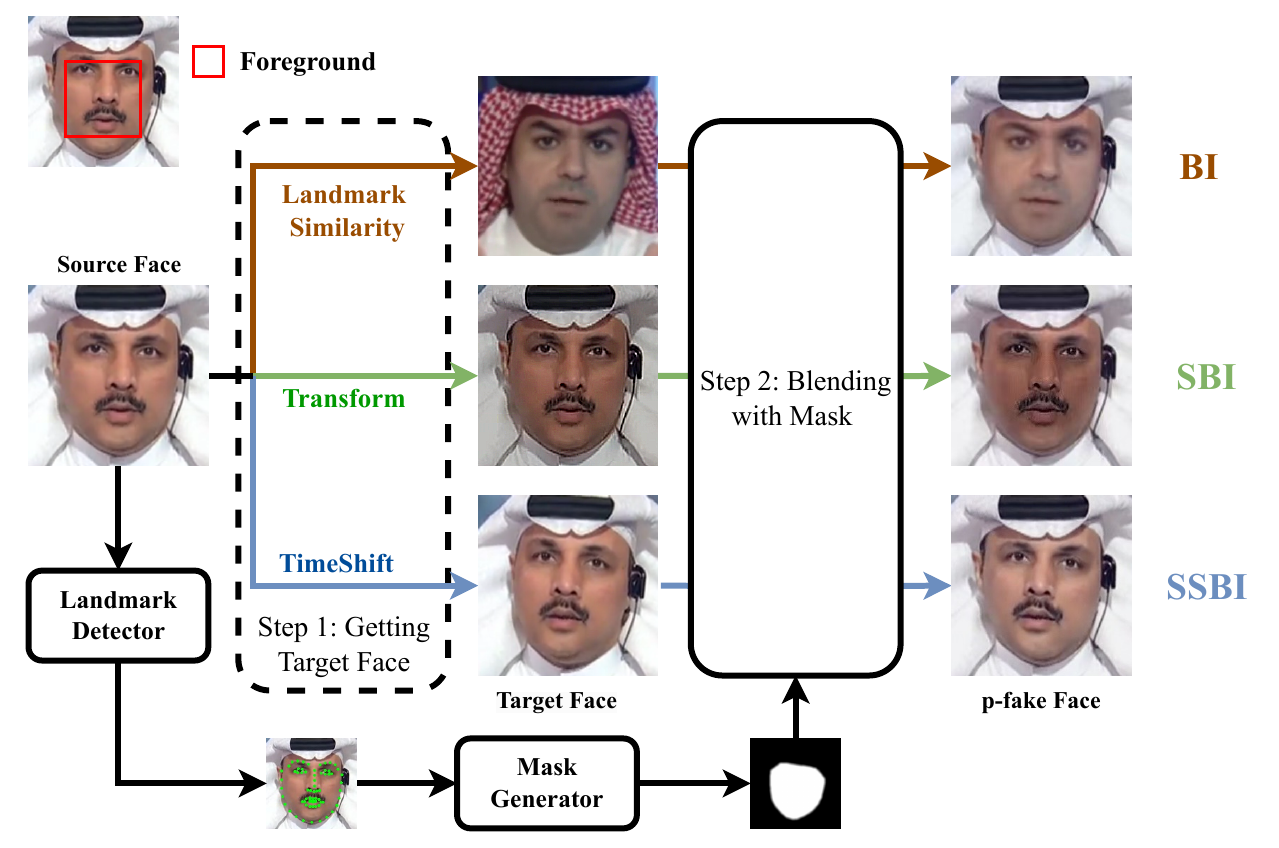}
	\caption{Overall pipeline of forgery augmentations.
	}
	\label{fig:fakeda}
\end{figure}

\subsection{Forgery Augmentation Operations}\label{subsec:fa}
Given a pristine source face image $x$, the forgery augmentation operations can simply be considered as modifying the foreground face region while keeping the background. To achieve this, forgery augmentation generally consists of the two steps, i.e., 1) get a target face $x_{t}$, 2) blending it with a mask $M$. To generate the blending mask $M$, we first extract the facial landmarks $l(x)$ by Dlib \cite{DlibmlMachineLearning2009king} and then apply random deformation and blurring on the convex hull, which is inspired by \cite{FaceXRayMore2020li,DetectingDeepfakesSelfBlended2022shiohara}. 
We obtain p-fake face $\hat{x}$ by:
\begin{equation}\label{eq:blend}
	\hat{x}= x_{t}\odot M+ x\odot (1-M)
\end{equation}
where $\odot$ specifies the element-wise multiplication.

For the selection of $x_{t}$, BI \cite{FaceXRayMore2020li}, the target face is get from different identities with top facial similarity to the source face. As for SBI \cite{DetectingDeepfakesSelfBlended2022shiohara}, the target face is get by source face itself with data transforms. 
However, the aformentioned works are dedicated to simulate the inconsistency in the spatial domain and thus cannot capture temporal inconsistencies across video frames, which is one of the important clues for identify deepfakes.

\noindent \textbf{Self-shifted Blending Image:}
We propose a novel forgery augmentation operation named \textit{Self-shifted Blending Image (SSBI)} to imitate temporal artifacts. The target face of SSBI is get from another frame on the same video. It can simply imitate temporal inconsistency between the foreground face and background in terms of face movements \cite{ExposingDeepFakes2019yang}.

The overall pipeline of the aforementioned forgery augmentations can be seen in Figure \ref{fig:fakeda}. 

\begin{algorithm}[t]  
	\caption{Policy-Controlled Forgery Augmentation $\mathcal{T}$}\label{alg:fda}
	\begin{algorithmic}[1]
		\REQUIRE Source face $x \sim \mathcal{D}^{r}_{tr}$, augmentation policy $p$
		\ENSURE Pseudo-fake face $\hat{x}$
		
		\STATE Get landmarks $l(x)$
		\STATE Sample one operation $j$ based on $p$
		
		\blue{\textit{\# Getting the target face $x_{t}$}} 
		\IF{$j$=1}
		\STATE Conduct BI \cite{FaceXRayMore2020li}: $x_{t}=argmin_{x_{t}\sim \mathcal{D}^{r}_{tr}} \left | l(x_{t})-l(x) \right |$
		\ELSIF{$j$=2}
		\STATE Conduct SBI \cite{DetectingDeepfakesSelfBlended2022shiohara}: $x_{t} = Transform(x)$
		\ELSIF{$j$=3}
		\STATE Conduct SSBI: $ x_{t} = TimeShift(x, rand(5,10))$
		\ENDIF
		
		\blue{\textit{\# Blending with mask}}
		\STATE $M=Deform(ConvexHull(l(x)))$
		\STATE Generate $\hat{x}$ by Equation (\ref{eq:blend})
		
	\end{algorithmic}
\end{algorithm}

\subsection{Dynamic Forgery Search}
Although the p-fake technology has been employed in some SOTA deepfake detection methods, the production of p-fake samples during the entire training period typically involves only a single forgery augmentation operation.
Taking inspiration from the success of automatic data augmentation techniques \cite{AutoAugmentLearningAugmentation2019cubuk,DifferentiableAutomaticData2020li,AdaAugLearningClass2022cheung,SurveyAutomatedData2023cheung} , we suggest that employing multipule forgery augmentation operations and dynamically adjusting their policy throughout the training process is better than relying solely on a fixed single forgery augmentation operation for training the general deepfake detector.
Thus, we propose to devise a strategy that dynamically selects optimal forgery augmentations based on evolving data and model states in different training stages.

To achieve this, we first define a policy-controlled forgery augmentation operator $\mathcal{T}(\cdot)$. 
Let $\mathbb{T}$ be a set of forgery augmentation operations where $\tau_j$ denotes the $j$-th operation. In this work, $\mathbb{T}$ only contains three forgery augmentation operations mentioned in Section \ref{subsec:fa} so that $|\mathbb{T}|=3$.
We formulate an forgery augmentation policy as the probability $p$ in applying multiple forgery augmentation operations. 
Here, $p$ is a probability vector with each entry: $p_j \in[0,1] ; \sum_{j=1}^{|\mathbb{T}|} p_j=1$.
As detailed in Algorithm \ref{alg:fda}, given an real face image $x$, we sample one operation according to an policy $p$ and get the p-fake sample by
\begin{equation}\label{eq:fa_op}
	\hat{x} = \mathcal{T}(x ; p)=\tau_j\left(x \right) ; j \sim p
\end{equation}
The generated p-fake sample $\hat{x}$ is labeled as fake.

Subsequently, we introduce the joint optimization of the deepfake detector and p-fake generation policy during the training.
We employ a feature extraction network $f_{\alpha}:\mathcal{D}\to \mathcal{Z}$ to map a data space to a latent space, a classification head $g_{\beta}:\mathcal{Z}\to \mathcal{Y}$ to map a latent space to a label space. $g_{\beta}  \circ f_{\alpha}$ can be regarded as a universal deepfake detector,where $\circ$ is the compositional operator.
We add a lightweight policy model $h_{\gamma}:\mathcal{Z}\to P$ to map a latent space to a probability space, where $p \in P$. 
The deepfake detector is optimized by minimizing the binary cross-entropy loss $L_{CE}$ on a training batch $\mathcal{B}_{tr}$.
The policy model is to search forgery augmentation policies applied to the training of the deepfake detector. Its optimization objective is to minimize $L_{CE}$ on search batch data, denoted as $\mathcal{B}_{sc}$. Herein, we sample the $\mathcal{B}_{sc}$ from the validation set $\mathcal{D}_{val}$.

Overall, the above objection can be formulated as a bi-level optimization problem\cite{InvestigatingBiLevelOptimization2022liu,AdaAugLearningClass2022cheung} as follow: 
\begin{equation}
	\begin{aligned}
		&\min _{\alpha, \beta} \mathcal{L}_{CE}\left(\alpha, \beta, \gamma^* ; \mathcal{B}_{tr}\right) \\
		s.t.  &\min _{\gamma} \mathcal{L}_{CE}\left(\alpha^*, \beta^*, \gamma; \mathcal{B}_{sc}\right) 
	\end{aligned}
\end{equation}
We solve it by executing the following optimization phase alternatively. 

\noindent \textbf{Optimization for Deepfake Detector.}
In this phase, given the training data $x$, the frozen policy model $h_{\gamma}$ generates the policy $p$. 
We get the augmented p-fake sample $\hat{x}$ using Equation (\ref{eq:fa_op}) and use them to train the detector model $g_{\beta}\circ f_{\alpha}$ by minimizing $L_{CE}$ on the processed mini-batch $\mathcal{B}_{tr}$.

\noindent \textbf{Optimization for Policy Model.}
In this phase, the weights of deepfake detector are frozen, and we aim to optimize $h_{\gamma}$ policy given the validation data.
However, we can not directly use back-propagation to optimize $\gamma$ because the sampling process of one forgery augmentation operation in $\mathcal{T}(x; p)$ is non-differentiable. Hence, back-propagation cannot compute the partial derivative w.r.t. the augmentation probability $p$. To address this problem, we relax the non-differentiable $\mathcal{T}(x; p)$ to be a differentiable operator, 
The augmented validation data are passed to the feature extraction network $f$ individually to get the latent representations, which are then summed based on their weights in the probability vector $p$. 
The forward pass can be relaxed as the mixed representation  and passed to $g$ for computing the predicted labels:
\begin{equation}\label{eq:se}
	\hat{y}=g_{\beta}\left(\sum_{j=1}^{|\mathbb{T}|} p_j \cdot f_{\alpha}(\tau_j(x))  \right) ; p=h_\gamma \circ f_{\alpha}(x)
\end{equation}
In this way, we can update $\gamma$ by minimizing $L_{CE}$ combined with back-propagation.

With the aforementioned updating rules, both the policy and detector models can be alternatively optimized. We set the search frequency $s$ to make that  Optimization for Policy Model is executed after every $s$ steps of Optimization for Deepfake Detector.

\section{Experiments}
\label{sec:exp}
\subsection{Experiment Settings}
\noindent \textbf{Datasets and pre-processing.}
Following most previous works, we mainly conducted experiments on the FaceForensics++ (FF++) \cite{FaceForensicsLearningDetect2019rossler} dataset. It contains 1000 Pristine (PT) videos (i.e., the real sample) and 5000 fake videos forged by five manipulation methods, i.e., Deepfakes (DF), Face2Face (F2F), FaceSwap (FS), NeuralTextures (NT) and FaceShifter (FSh). Besides, FF++ provides three quality levels in compression for these videos: raw, high-quality (HQ) and low-quality (LQ). The \textbf{HQ version of FF++} is adopted by default in this paper. If any deviation from this default, it will be explicitly stated. The samples were split into disjoint training, validation, and testing sets at the video level follows the official protocol. 

To demonstrate the performance of CDFA in cross-dataset settings, four additional datasets are adopted, i.e., Celeb-DF-v2 (CDF) \cite{CelebDFLargeScaleChallenging2020li}, DeepFake Detection Challenge preview (DFDCP) \cite{DeepfakeDetectionChallenge2019dolhansky} and DeepFake Detection Challenge public (DFDC)\cite{DeepFakeDetectionChallenge2020dolhansky} and WildDeepfake(Wild) \cite{WildDeepfakeChallengingRealWorld2020zi}. See the supplementary material for more details.

\noindent \textbf{Implementation details.}
We use SwinTransformerV2-Base (Swin) \cite{SwinTransformerV22022liu} as the backbone network $f_{\alpha}$, and the parameters are initialized by the weights pre-trained on the ImageNet. We implemented $h_{\gamma}$ using three MLP layers with random initialization, and the softmax operation is applied to the output of $h_{\gamma}$ to get the probabilities.
We use the Adam optimizer for both the two networks with a cosine learning rate scheduler initiate with 0.0001. We set the total training epoch $T=50$ and the searching frequency as $s=10$. 
See the supplementary material for more details.

\noindent \textbf{Evaluation Metrics.}
In this work, we mainly report the area under the ROC curve (AUC) to compare with prior works. The video-level results are obtained by averaging predictions over each frame on an evaluated video.

\begin{table}[t]
	\centering
	\caption{Video-level (top) and frame-level (bottom) AUC(\%)  of cross-datasets performances compared with SOTA methods. The best results are highlighted.}
	\label{tab:cross-datasets}
	\begin{tabular}{ccccccc} 
		\toprule
		Method                                                                            & Backbone  & Data & CDF            & Wild           & DFDCP          & DFDC            \\ 
		\midrule
		TALL\cite{TALLThumbnailLayout2023xu}                      & Swin      & HQ   & 90.79          & -              & -              & 76.78           \\
		SeeABLE\cite{SeeABLESoftDiscrepancies2023larue}            & ENb4      & HQ   & 87.30          & -              & 86.30          & 75.90           \\
		CADDM\cite{ImplicitIdentityLeakage2023dong}                   & ENb4      & HQ   & 93.88          & -              & -              & 73.85           \\
		AUNet\cite{AUNetLearningRelations2023bai}                & Xcep      & HQ   & 92.77          & -              & 86.16          & 73.82           \\
		LTTD\cite{DelvingSequentialPatches2022guana}                  & Designed & HQ   & 89.30          & -              & -              & 80.40           \\
		CD-NET\cite{AdaptiveFaceForgery2022song}                     & Xcep      & HQ   & 88.50          & -              & -              & 77.00           \\
		DCL\cite{DualContrastiveLearning2022sun}                       & ENb4      & HQ   & 82.30          & 71.14          & -              & 76.71           \\
		PCL+I2G\cite{LearningSelfConsistencyDeepfake2021zhao}                     & Res34      & HQ   & 90.03          & -              & 74.37              & 67.52           \\
		\midrule
		Ours                                                                              & Swin      & HQ   & \textbf{97.22} & \textbf{84.45} & \textbf{97.03} & \textbf{83.84}  \\
		Ours                                                                              & Swin      & LQ   & 94.63          & 84.05          & 96.60          & 81.16           \\ 
		\midrule
		LSDA\cite{TranscendingForgerySpecificity2024yan}                            & ENb4      & HQ   & 83.00          & -              & 81.50          & 73.60           \\
		UCF\cite{UCFUncoveringCommon2023yan}                            & Xcep      & HQ   & 75.27          & -              & 75.94          & 71.91           \\
		SFDG\cite{DynamicGraphLearning2023wang}                        &  ENb4         & LQ   & 75.83          & 69.27          & -              & 73.64           \\
		NoiseDF\cite{NoiseBasedDeepfake2023wang}                       &  Designed         & HQ   & 75.89          & -              & -              & 63.89           \\
		OST\cite{OSTImprovingGeneralization2022chena}                   &   Xcep        & HQ   & 74.80          & -              & -              & \textbf{83.30}  \\
		UIA-ViT\cite{UIAViTUnsupervisedInconsistencyAware2022zhuang}    & ViT-B     & HQ   & 82.41          & -              & 75.80          & -               \\
		SLADD\cite{SelfSupervisedLearningAdversarial2022chen}          & Xcep      & HQ   & 79.70          & -              & -              & 77.20           \\
		RECCE\cite{EndtoEndReconstructionClassificationLearning2022cao} & Xcep         & LQ   & 68.71          & 64.31          & -              & 69.06           \\
		PEL\cite{ExploitingFineGrainedFace2022gu}                      & ENb4      & LQ   &  69.18          & 67.39          & -              & 63.31           \\ 
		\midrule
		Ours                                                                              & Swin      & HQ   & \textbf{91.96} & \textbf{81.34} & \textbf{93.30} & 81.45           \\
		Ours                                                                              & Swin      & LQ   & 89.88          & 80.99          & 92.65          & 78.67           \\
		\bottomrule
	\end{tabular}
\end{table}

\subsection{Generalization Comparisons}
To comprehensively evaluate the generalizability of our method, we compare the performances of cross-datasets and cross-manipulation evaluations with several SOTA methods published in the past three years.

\noindent \textbf{Cross-datasets evaluations.}
The cross-datasets evaluation is still a challenging task because the unknown domain gap between the training and testing datasets can be caused by different source data, forgery methods, and/or post-processing. In this part, we evaluate the generalization performances in a cross-dataset setting. Specifically, our models were trained on the FF++ (only containing DF, F2F, FS, and NT subsets for fair comparisons) and tested on other datasets.
The experimental results in terms of frame-level and video-level AUC are shown in Table \ref{tab:cross-datasets}. We can observe that our method outperforms the best competition in terms of video-level evaluations. For frame-level evaluations, our method still outperforms most of the SOTA competitors regardless it is trained on the HQ or LQ version of FF++.
For instance, our approach surpasses TALL\cite{TALLThumbnailLayout2023xu}, which also employs Swin as its backbone network, by around 7\% when testing on CDF and DFDC.
We can also see that our method obtains a lower frame-level AUC when testing on DFDC compared to OST\cite{OSTImprovingGeneralization2022chena}. 
One possible explanation is that OST introduces a test-time adaptation strategy that adapts the model with domain knowledge of testing data before evaluation. This trick facilitates for evaluating large-scale unseen data such as DFDC. However, our method never introduces knowledge from testing data during the training.

\noindent \textbf{Backbone impact.}
In Table \ref{tab:bb_cross}, we evaluated the performances of CDFA when employing different backbone architectures $f_{\alpha}$, i.e., Xception (Xcep) \cite{XceptionDeepLearning2017chollet}, EfficientNetb4 (ENb4) \cite{EfficientNetRethinkingModel2019tan} and Swin \cite{SwinTransformerV22022liu}. 
We observe that our CDFA can significantly improve the generalization performances of all evaluated models (at least 13\% on average).
In conjunction with the results in Table \ref{tab:cross-datasets}, CDFA still outperforms the SOTA competitors even with the same backbones (e.g., the comparison of SeeABLE and ENb4+CDFA). 
We also find that larger and more powerful encoders lead to better generalization in general when equipping CDFA.
These results suggest that CDFA is applicable to different backbone models and is expected to further benefit from future developments in model topologies.

\begin{table}[t]
	\centering
	\begin{minipage}{0.49\textwidth}
		\centering
		\caption{Video-level AUC(\%) on cross-dataset evaluations (trained on FF++) performances. The best results are highlighted.}
		\resizebox{\textwidth}{!}{
		\begin{tabular}{cccccc} 
			\toprule
			Methods & CDF            & Wild           & DFDCP          & DFDC         & Avg    \\ 
			\midrule
			Xcep    & 69.99          & 60.06          & 80.93          & 65.85         & 69.21   \\
			+ CDFA  & \textbf{93.24} & \textbf{78.06} & \textbf{86.43} & \textbf{77.06} & \textbf{84.56}  \\ 
			\midrule
			ENb4    & 74.50          & 61.45          & 82.28          & 68.15          & 71.60  \\
			+ CDFA  & \textbf{95.81} & \textbf{78.76} & \textbf{86.66} & \textbf{78.38}  & \textbf{85.25} \\ 
			\midrule
			Swin  & 73.53          & 71.56          & 89.37          & 71.30          & 76.44  \\
			+ CDFA  & \textbf{97.22} & \textbf{84.45} & \textbf{97.03} & \textbf{83.84} & \textbf{91.63}  \\
			\bottomrule
		\end{tabular}
	}
		\label{tab:bb_cross}
	\end{minipage}
	\hfill
	\begin{minipage}{0.49\textwidth}
		\centering
		\caption{Video-level AUC(\%) on cross-manipulation evaluations. Trained on FF++/DF. The best cross-manipulation results are highlighted.}
		\resizebox{\textwidth}{!}{
		\begin{tabular}{cccccc} 
			\toprule
			Method   & DF    & F2F            & FS             & NT             & FSh             \\ 
			\midrule
			CADDM \cite{ImplicitIdentityLeakage2023dong}    & 100   & 83.94          & 58.33          & 68.98          & -               \\
			DCL \cite{DualContrastiveLearning2022sun} & 99.98 & 77.13              & 61.01          & 75.01              & -           \\
			\midrule
			Xcep & 100     & 73.60          & 53.73          & 71.53          & 71.62           \\
			+ CDFA    & 99.54 & \textbf{85.93} & \textbf{90.04} & \textbf{82.23} & \textbf{77.81}  \\
			\midrule
			ENb4    & 100   & 71.23          & 47.32          & 70.31          & 75.71           \\
			+ CDFA    & 99.65 & \textbf{88.51} & \textbf{91.84} & \textbf{82.14} & \textbf{84.57}  \\
			\midrule
			Swin     & 100   & 67.43          & 56.74          & 78.74          & 70.87           \\
			+ CDFA    & 99.90 & \textbf{87.44} & \textbf{90.64} & \textbf{86.27} & \textbf{76.64}  \\
			\bottomrule
		\end{tabular}
	}
		\label{tab:cross-manipulation}
	\end{minipage}
\end{table}

\noindent \textbf{Cross manipulation evaluations.}
In real detection situations, the defenders generally are not aware of the attacker’s forgery methods. For this reason, it is important to verify the model generalization to various forgery methods. We conducted the cross-manipulation experiment on FF++, all models were trained on the DF subset and tested on the remaining four manipulations. 
More results on other subsets are given in the Supplementary Material.
We evaluate the effect of different backbone architectures $f_{\alpha}$ equipped with the proposed CDFA.
As shown in Table \ref{tab:cross-manipulation}, we can observe that our CDFA can improve cross-manipulation performances significantly regardless of the types of backbones.
In addition, the backbone models trained with the CDFA approach outperform the SOTA competitors (i.e, CADDM \cite{ImplicitIdentityLeakage2023dong} and DCL \cite{DualContrastiveLearning2022sun}) by a considerable margin on average.
These results highlight the effectiveness of CDFA in combating emerging unseen forgery methods.

\subsection{Ablation Studies}\label{subsec:abl}  
In this part, we perform several ablations to better understand the contributions of each component in the proposed CDFA, including consists of fake data during the training, monotonic curriculum and dynamic forgery search.
We evaluated several variants of the proposed CDFA (Trained on FF++/DF) and summarized the results in Table \ref{tab:abl}.

\begin{table}[t]
	\centering
	\caption{Video-level AUC(\%) performances for ablation studies. }
	\label{tab:abl}
	\begin{tabular}{cccccccccc} 
		\hline
		\multirow{2}{*}{Variant} & \multicolumn{2}{c}{Fake data} & \multicolumn{2}{c}{Strategies} & \multirow{2}{*}{CDF} & \multirow{2}{*}{Wild} & \multirow{2}{*}{DFDCP} & \multirow{2}{*}{DFDC} & \multirow{2}{*}{Avg}  \\ 
		\cmidrule(lr){2-3} \cmidrule(lr){4-5}
		& o-fake  & p-fake              & MC      & DFS                           &                      &                       &                        &                       &                       \\ 
		\hline
		1                        & -       & BI                  & -       & -                             &  88.40                     &     82.71           &        87.86                &   74.55                    &        83.38               \\
		2                        & -       & SBI                 & -       & -                             &      91.31                &           76.14            &       91.57                 &         70.14              &           82.29            \\
		3                        & -       & SSBI                & -       & -                             &         90.60             &       83.31                &          95.98              &           76.76            &          86.67             \\
		4                        & -       & ALL                 & -       & -                       & 95.03                & 77.39                 & 90.49                  & 77.41                 & 85.08                 \\
		5                        & -       & ALL                 & -       & $\surd$                                &         95.84             &          82.47             &          93.18              &            82.27           &           88.44            \\
		6                        & $\surd$ & BI                  & $\surd$ & -                             & 93.34                & 80.16                 & 89.64                  & 78.37                 & 85.38                 \\
		7                        & $\surd$ & SBI                 & $\surd$ & -                             & 94.41                & 75.37                 & 88.36                  & 73.54                 & 82.92                 \\
		8                        & $\surd$ & SSBI                & $\surd$ & -                             & 94.75                & 83.30                 & 95.52                  & 83.95                 & 89.38                 \\
		9                        & $\surd$ & ALL                 & $\surd$ & -                             & 97.26                & 85.22                 & 96.46                  & 82.20                 & 90.29                 \\
		10                       & $\surd$ & ALL                 & -       & -                             & 96.31                & 81.57                 & 95.61                  & 83.79                 & 89.32                 \\
		11                       & $\surd$ & ALL                 & -       & $\surd$                       & 97.06                & 84.72                 & 95.73                  & 86.37                 & 90.97                 \\
		\hline
		CDFA                     & $\surd$       & ALL                 & $\surd$ & $\surd$                       & \textbf{97.94}                & \textbf{86.72}                 & \textbf{97.63}                  & \textbf{87.16}                 & \textbf{92.36}                 \\
		\hline
	\end{tabular}
\end{table}

\noindent \textbf{Effects of the forgery augmentation operations.}
From the comparison among \textit{Variant 1}, \textit{Variant 2}, and \textit{Variant 3}, where the fake part of the training data only contained p-fake samples generated by single forgery augmentation operation, we observe that our proposed SSBI performs better compared to BI and SBI in most results. 
Similar phenomena also appear in the comparison of \textit{Variant 6}, \textbf{7}, and \textit{Variant 8}.
These results highlight the effectiveness of the proposed SSBI which compensates the deficiency in simulating temporal artifacts.

\noindent \textbf{Effects of monotonic curriculum.}
From the comparative analysis of \textit{Variant 1} with \textit{Variant 6}, \textit{Variant 2} with \textit{Variant 7}, \textit{Variant 3} with \textit{Variant 8}, and \textit{Variant 4} with \textit{Variant 9}, we note that using of o-fake samples through MC strategies can improve the performances in most cases compared to only training with p-fake samples. Furthermore, the comparison between \textit{Variant 9} and \textit{Variant 10} indicates that when using o-fake, using MC strategies performs better than simply mixing them into p-fake samples.
It shows the effectiveness of the idea behind MC, which is to use o-fake at the beginning and gradually introduce more p-fake during the training.

\noindent \textbf{Effects of dynamic forgery search.}
The comparative analysis of the \textit{Variant 3}, \textit{Variant 4} and \textit{Variant 5} indicates that utilizing all forgery augmentation operations with fixed probability can not improve the performances compared with using SSBI alone, while the introduction of DFS leads to significant improvements.
Furthermore, from the comparative analysis of the \textit{Variant 5}, \textit{Variant 11} and CDFA, we can find that the improvements brought by DFS can be further increased with the guidance of MC.
It shows the effectiveness of DFS which dynamically searches the optimal forgery augmentation policy during the training.

\begin{figure}[t]
	\centering
	\subfloat[Our CDFA]{\includegraphics[width=0.35\linewidth]{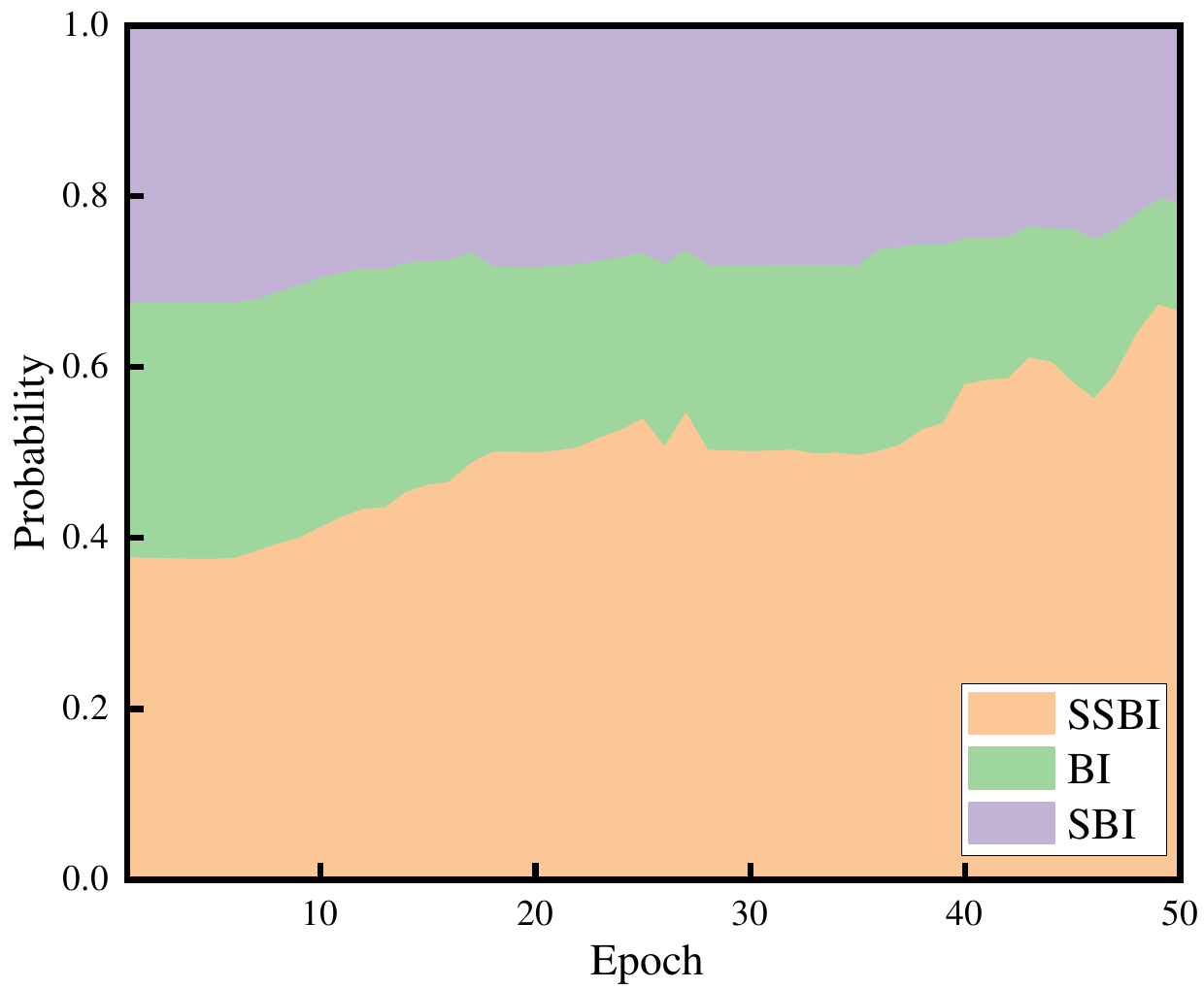} \label{subfig:cdfa}}
	\subfloat[Variant 11 in Table \ref{tab:abl}]{\includegraphics[width=0.35\linewidth]{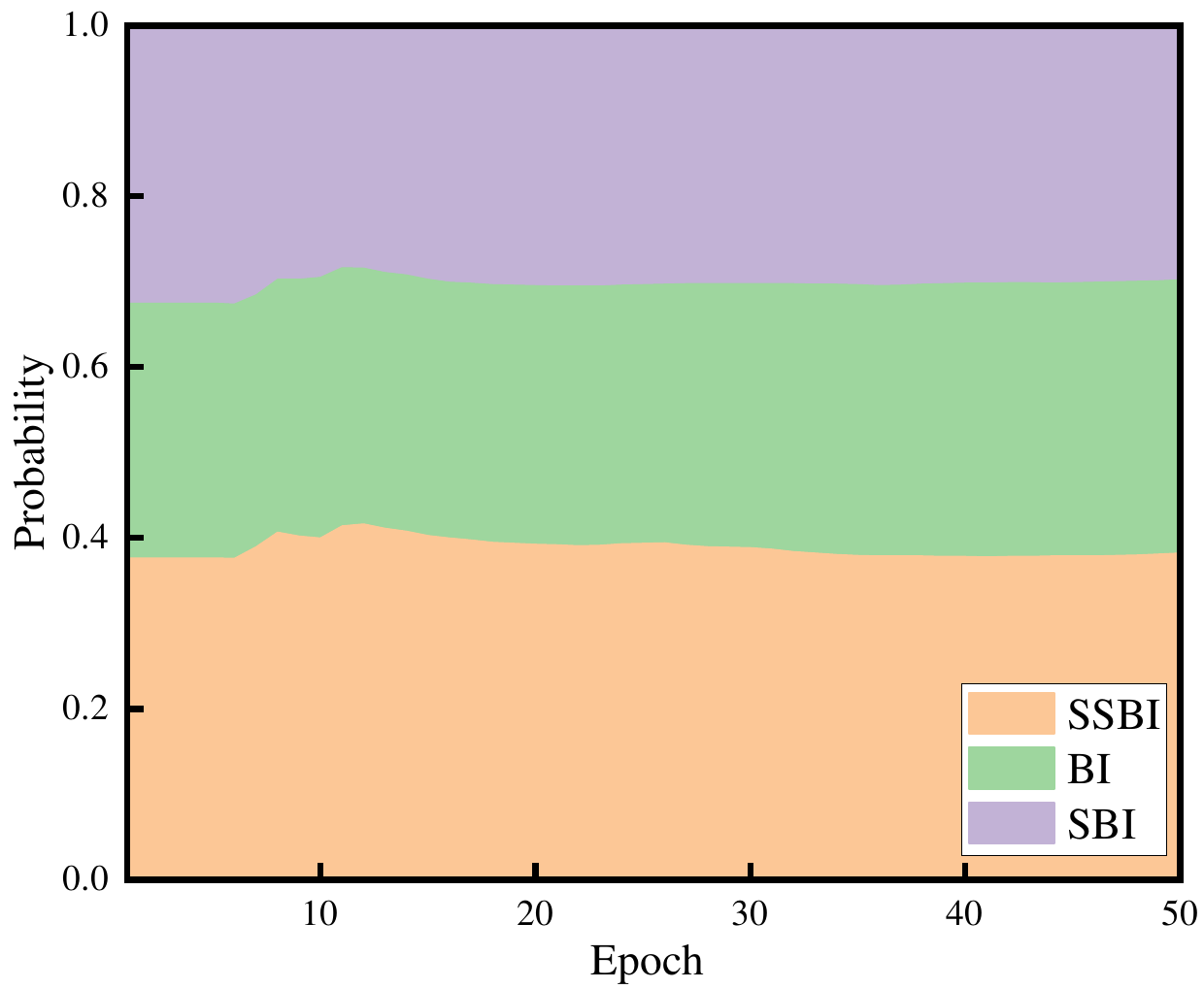} \label{subfig:v11}}
	\caption{Evolution of policies searched by DFS for two variants method: \textbf{(a) }the proposed CDFA, \textbf{(b)} Variant 11 in Table \ref{tab:abl}. 
		The policy of one epoch is obtained by averaging the policies searched at all steps over the epoch.}
	\label{fig:policy}
\end{figure}
\subsection{Visualizations and Analysis}
\noindent \textbf{Analysis of searched policies.}
In this part, we depict and analyze the evolution of policies searched by DFS throughout the training process. 
For CDFA, as shown in Figure \ref{subfig:cdfa}, we observe a progressive rise in the probability of SSBI as training proceeds, whereas the probabilities of BI and SBI gradually decline. 
Such a phenomenon suggests that the model places increasing importance on p-fake samples generated by SSBI as the training proceeds.
This observed evolution of policies also aligns with the results in Table \ref{tab:abl}, i.e., the performances of using SSBI alone surpass that of BI and SBI alone. It further emphasizes the effectiveness of the proposed SSBI. 
Moreover, from Figure \ref{subfig:v11}, it is apparent that maintaining a constant proportion of o-fake samples throughout the training does not leverage the full potential of DFS. This constancy may cause DFS to become less dynamic, failing to adaptively adjust the training strategy to optimize deepfake detection effectively.
These results highlight the importance of the guidance introduced by MC strategy.

\begin{figure}[t]
	\centering
	\begin{minipage}[t]{0.45\textwidth} 
		\centering 
		\includegraphics[width=0.8\linewidth]{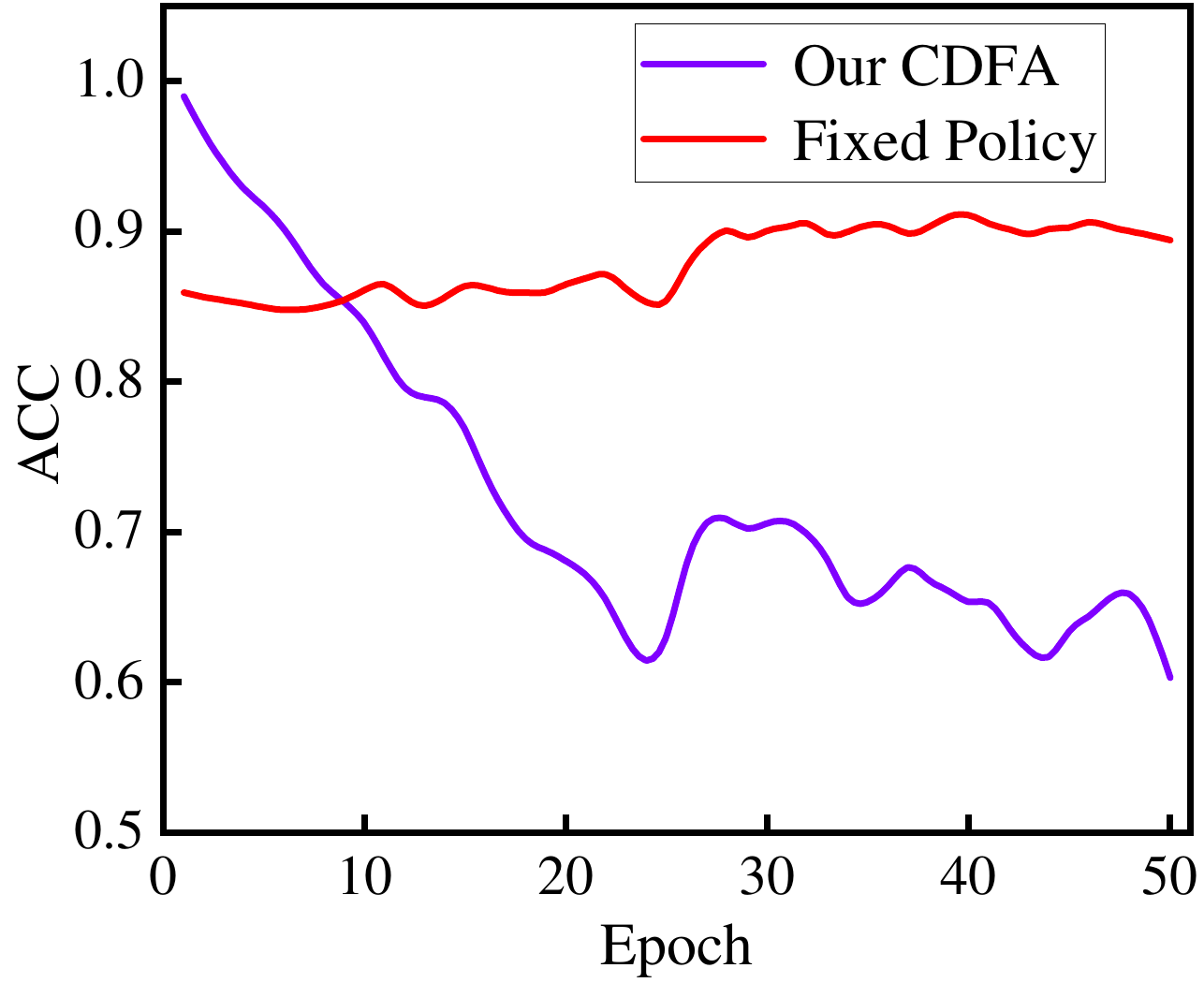}
		\caption{Assessing accuracy during the training.}
		\label{fig:hardness}
	\end{minipage}
	\hfill 
	\begin{minipage}[t]{0.45\textwidth}
		\centering
		\includegraphics[width=0.7\linewidth]{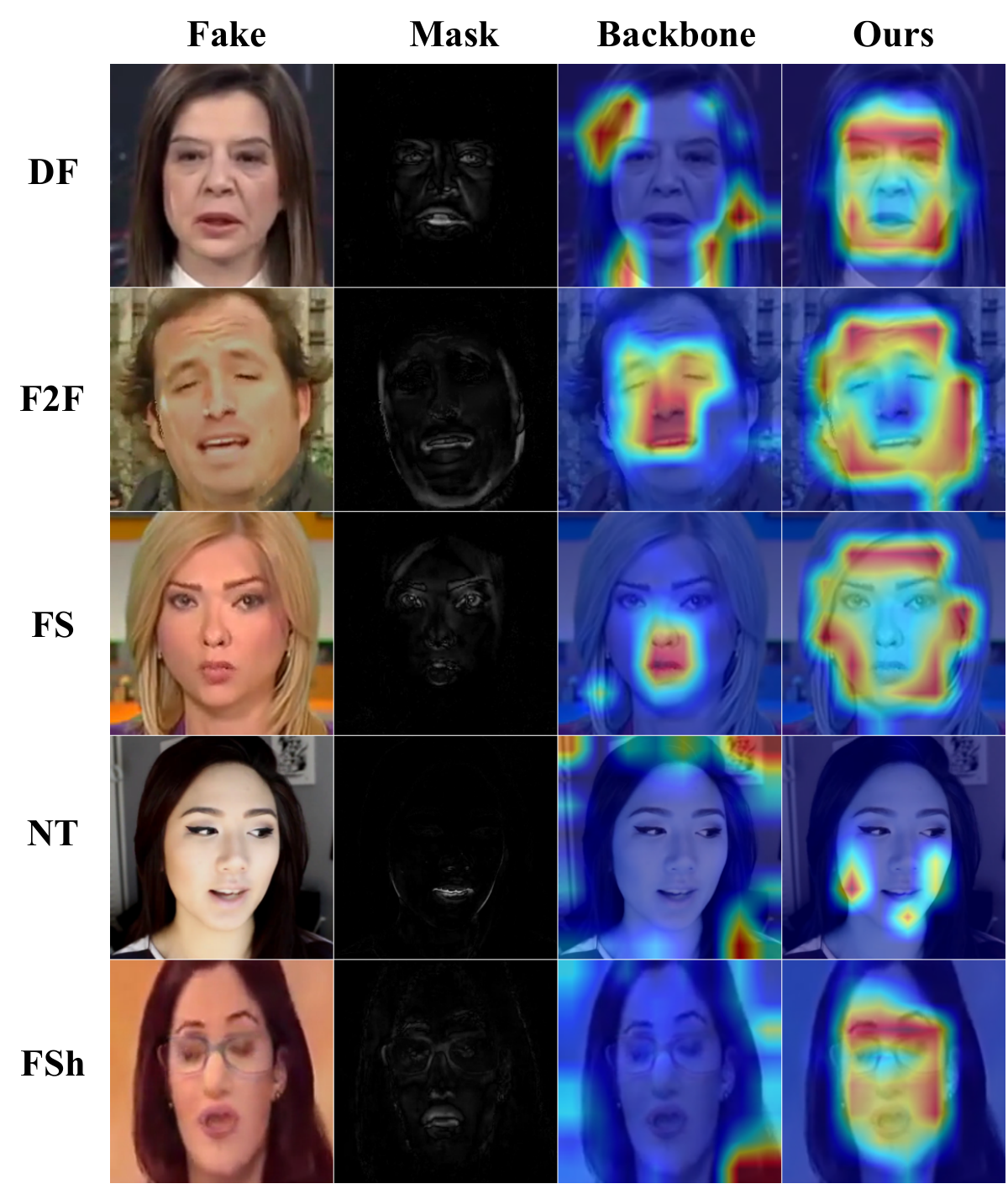}
		\caption{GradCAM visualizations of the backbone and our proposed CDFA.}
		\label{fig:gradcam}
	\end{minipage}
\end{figure}

\noindent \textbf{Analysis of fake samples in training process.}
In this part, we study the properties of fake samples during the training.
Specifically, we employed a baseline deepfake detector (i.e., Swin \cite{SwinTransformerV22022liu}) as an assessment model. 
We first train the assessment model on FF++ and then fix it to assess the fake samples utilized in each training epoch. 
We depict the assessing accuracy of our CDFA and fixed policy of o-fake and p-fake samples  (i.e., \textit{Variant 10} in Table \ref{tab:abl}) in Figure \ref{fig:hardness}.
It can be observed that assessment accuracy decreases significantly in the early stage of training, while it fluctuates in the later stages.
This observation suggests that our CDFA gradually increases the difficulty of fake samples via MC in the early stages of training while maintaining their diversity via DFS in the later stages of training when p-fake dominates the fake samples.
It reveals that the deepfake detector can learn more general forgery representations by gradually focusing on hard fake samples with diversity during the training.

\noindent \textbf{Visualizations of CAM.}
To intuitively demonstrate which patterns learned with our proposed CDFA, we compare the GradCAM \cite{GradCAMVisualExplanations2017selvaraju} visualizations between the backbone model (i.e., Swin) trained with and without CDFA. As shown in Figure \ref{fig:gradcam}, the backbone trained without CDFA tend to only capture method-specific artifacts. Our method identifies the forgery faces by focusing on the general artifacts (e.g., the blending traces on the boundary between the background and foreground face) with the help of CDFA. Based on the results of our quantitative experiments (Table \ref{tab:bb_cross}), we believe that paying attention to the inconsistency between the background and facial parts can improve the generalization ability of deepfake detectors.

\section{Conclusions}
\label{sec:con}
In this work, we present CDFA, which aims to improve the generalization performances of deepfake detectors by dynamically adjusting the composition of fake samples during the training.
First, we propose a monotonic curriculum that progressively increases the proportion of p-fake samples as training proceeds. 
Second, we propose a dynamic forgery searching strategy to conduct the optimal forgery augmentation operation for each image varying between training stages. In addition, we propose a novel forgery augmentation scheme named SSBI to simply imitate the temporal inconsistency of deepfake generation.
Comprehensive experiments show that CDFA can significantly improve the performances of various naive deepfake detectors in a plug-and-play way, and make them attain superior performances over the existing methods in several cross-datasets and cross-manipulations benchmarks. 

\noindent \textbf{Ethic Statement.} 
All face images used in this paper were obtained from public datasets. There is no violation of personal privacy while conducting experiments in this work.

\section*{Acknowledgments}
This work was supported in part by National Natural Science Foundation of China (Grant U23B2022, U22B2047, U22A2030), Guangdong Provincial Key Laboratory (Grant 2023B1212060076) and Guangdong Major Project of Basic and Applied Basic Research (Grant No. 2023B0303000010). The work was also supported in part by China Postdoctoral Science Foundation under Grant 2021TQ0314 and Grant 2021M703036.
%
%
\bibliographystyle{splncs04}
\bibliography{11581-refs}

\clearpage
\setcounter{page}{1}
\setcounter{table}{0}
\setcounter{figure}{0} 
\appendix
\renewcommand\thefigure{\Alph{section}\arabic{figure}}
\renewcommand\thetable{\Alph{section}\arabic{table}}     

\renewcommand{\thesection}{\Alph{section}}
\title{Fake It till You Make It: Curricular Dynamic Forgery Augmentations towards General Deepfake Detection-Supplementary Material} 

\titlerunning{Curricular Dynamic Forgery Augmentations for Deepfake Detection-Supp}


\author{Yuzhen Lin \inst{1}\orcidlink{0000-0001-7788-2054} \and
	Wentang Song \inst{1}\orcidlink{0000-0003-3750-9516} \and
	Bin Li \inst{1}\orcidlink{0000-0002-2613-5451} \and
	Yuezun Li\inst{2}\orcidlink{0000-0001-9299-1945} \and
	Jiangqun Ni\inst{3}\orcidlink{0000-0002-7520-9031} \and
	Han Chen\inst{1}\orcidlink{0000-0002-9439-9133} \and
	Qiushi Li\inst{1}\orcidlink{0000-0002-4976-3346}
}

\authorrunning{Lin et al.}

\institute{Guangdong Provincial Key Laboratory of Intelligent Information Processing,  
	Shenzhen Key Laboratory of Media Security, 
	SZU-AFS Joint Innovation Center for AI Technology, 
	Shenzhen University, Shenzhen, China \\
	\email{\{linyuzhen2020, 2018132120, 2016130205, 1800271017\}@email.szu.edu.cn; \Envelope~libin@szu.edu.cn;}
	\and
	College of Computer Science and Technology, Ocean University of China, Qingdao, China; 
	\email{liyuezun@ouc.edu.cn}
	\and
	School of Cyber Science and Technology, Sun Yat-Sen University, and Department of New Networks, Peng Cheng Laboratory, Shenzhen, China \\
	\email{issjqni@mail.sysu.edu.cn}
}

\maketitle

\section{More Implementation Details}
Our proposed approach is implemented by PyTorch on a workstation equipped with one NVIDIA Tesla A100 GPU (40GB memory).
To provide further clarity on our method, we present the pipeline of the proposed CDFA in Algorithm \ref{alg:cdfa}, which outline the detailed steps.  
The hyper-parameters are set as $T_w=5, b=64$.

As for pre-processing, we utilized MTCNN to detect and crop the face regions (enlarged by a factor of 1.3) from each video frame, and resized the them to 256 $\times$ 256.

\section{More Details of Experimental Settings}
\label{suppsec:imp}
%
\subsection{More Details of Datasets}
We conduct evaluations on widely-used datasets and follow previous settings used in their corresponding datasets and compare with other methods respectively. More details on these datasets are described below.
\begin{itemize}
	\item \textbf{CelebDF (CDF)} \cite{CelebDFLargeScaleChallenging2020li} contains 590 real videos of 59 celebrities and corresponding 5639 high-quality fake videos generated by an improved forgery method. We use the stand test set consisting of 518 videos for our experiments.
	\item \textbf{DeepFake Detection Challenge Preview (DFDCP)} \cite{DeepfakeDetectionChallenge2019dolhansky} is generated by two kinds of synthesis methods on 1131 original videos. We use all 5250 videos for our experiments.
	\item \textbf{DeepFake Detection Challenge (DFDC)} \cite{DeepFakeDetectionChallenge2020dolhansky} is widely acknowledged as the most challenging dataset due to containing many manipulation methods and perturbation noises. We use the public test set consisting of 5000 videos for our experiments.
	\item \textbf{WildDeepfake (Wild)} \cite{WildDeepfakeChallengingRealWorld2020zi} contains 3805 real face sequences and 3509 fake face sequences collected from Internet. Thus, it has a variety of synthesis methods and backgrounds, as well as character identities. We use the stand test set consisting of 806 sequences for our experiments.
\end{itemize}

\subsection{More Details of Compared SOTA Methods}
In this work, we compare our method with several SOTA methods published in recent three years, including: TALL \cite{TALLThumbnailLayout2023xu}, SeeABLE \cite{SeeABLESoftDiscrepancies2023larue}, CADDM \cite{ImplicitIdentityLeakage2023dong}, AUNet \cite{AUNetLearningRelations2023bai}, LTTD \cite{DelvingSequentialPatches2022guana}, CD-NET \cite{AdaptiveFaceForgery2022song}, DCL \cite{DualContrastiveLearning2022sun}, PCL+I2G\cite{LearningSelfConsistencyDeepfake2021zhao},  UCF \cite{UCFUncoveringCommon2023yan}, SFDG \cite{DynamicGraphLearning2023wang}, NoiseDF \cite{NoiseBasedDeepfake2023wang}, OST \cite{OSTImprovingGeneralization2022chena}, UIA-ViT \cite{UIAViTUnsupervisedInconsistencyAware2022zhuang}, SLADD \cite{SelfSupervisedLearningAdversarial2022chen}, RECCE \cite{EndtoEndReconstructionClassificationLearning2022cao} and PEL \cite{ExploitingFineGrainedFace2022gu}.

As most previous works do, we refer to the reported results from the original papers of the aforementioned competitors.

\begin{algorithm}[t]  
	\caption{Curricular Dynamic Forgery Augmentation}\label{alg:cdfa}
	\begin{algorithmic}[1]
		\REQUIRE Training set $\mathcal{D}^{r}_{tr}, \mathcal{D}^{f}_{tr}$, Real part of validation set $\mathcal{D}^{r}_{val}$, epoch number $T$, warm-up epoch $T_w$, Batch size $b$, Searching frequency $s$.
		\ENSURE Model parameters $\alpha,\beta,\gamma$
		
		\FOR{$t=0$ to $T$}
		\FOR{$step=0$ to $D_{tr}/ b$}
		\STATE Sample $\mathcal{B}^{r}_{tr}\subseteq \mathcal{D}^{r}_{tr}, |\mathcal{B}^{r}_{tr}|=b/2$
		
		\blue{\textit{\# Monotonic Curriculum:}}
		\STATE Compute $q(t)$ with Equation (\ref{eq:mc}), $n_{of}, n_{pf}$ with Equation (\ref{eq:mcn})
		\STATE Sample $\mathcal{B}^{of}_{tr}\subseteq \mathcal{D}^{f}_{tr}, |\mathcal{B}^{of}_{tr}|=n_{of}$
		
		\blue{\textit{\# Optimization for Deepfake Detector:}}
		\STATE Apply policy model $h_{\gamma} \circ f_{\alpha}$ on random $n_{pf}$ samples in $\mathcal{D}^{r}_{tr}$ to get $\mathcal{B}^{pf}_{tr}$ with Equation (\ref{eq:fa_op})
		\STATE Construct $\mathcal{B}_{tr}$ with Equation (\ref{eq:batch})
		\STATE Update $g_{\beta}\circ f_{\alpha}$ on the processed $\mathcal{B}_{tr}$
		
		\blue{\textit{\# Optimization for Policy Model:}}
		\IF{$t>T_w$ and $step \mod s=0$}
		\STATE Sample $\mathcal{B}_{sc} \subseteq \mathcal{D}^{r}_{val}, |\mathcal{B}^{r}_{sc}|=b/2$
		\STATE Apply Equation (\ref{eq:se}) for each sample $x \in \mathcal{B}^{r}_{sc}$ to generate $\mathcal{B}^{pf}_{sc}$
		\STATE Update policy network $h_{\gamma}$ on $\mathcal{B}_{sc}=\mathcal{B}^{r}_{sc} \cup \mathcal{B}^{pf}_{sc}$.
		\ENDIF
		\ENDFOR
		\ENDFOR

	\end{algorithmic}
\end{algorithm}

\section{Additional Experimental Results}

\begin{table}
	\centering
	\caption{More video-level AUC(\%) results on cross-manipulation evaluations. The best results in cross-manipulation settings are highlighted.}
	\label{tab:supp-cross-man}
	\begin{tabular}{ccccccc} 
		\toprule
		\multicolumn{1}{l}{Training Data} & Method & DF    & F2F            & FS             & NT             & FSh             \\ 
		\midrule
		\multirow{4}{*}{F2F}              & CADDM \cite{ImplicitIdentityLeakage2023dong}  & \textbf{99.88}   & 99.97          & 79.40          & 82.38          & -               \\
		& DCL \cite{DualContrastiveLearning2022sun}    & 91.91 & 99.21          & 59.58          & 66.67          & -               \\
		& Swin   & 87.95   & 99.73          & 47.16          & 54.95          & 55.84           \\
		& Ours & 99.28 & 99.20 & \textbf{96.19} & \textbf{82.81} & \textbf{74.12}  \\
		\midrule
		\multirow{4}{*}{FS}              & CADDM \cite{ImplicitIdentityLeakage2023dong}  & 93.42   & 74.00          & 99.92          & 49.86          & -               \\
		& DCL \cite{DualContrastiveLearning2022sun}   & 74.80 & 69.75          & 99.90          & 52.60          & -               \\
		& Swin   & 53.62   & 72.15          & 100          & 43.36          & 46.01          \\
		& Ours & \textbf{99.11} & \textbf{93.02} & 99.93 &\textbf{75.72} & \textbf{77.81}  \\
		\midrule
		\multirow{4}{*}{NT}              & CADDM \cite{ImplicitIdentityLeakage2023dong}  & \textbf{100}   & \textbf{97.93}         & 86.76          & 99.46         & -               \\
		& DCL \cite{DualContrastiveLearning2022sun}   & 91.23 & 52.13          & 79.31          & 98.97          & -               \\
		& Swin   & 93.43   & 68.82          & 42.46          & 98.15          & 68.28           \\
		& Ours & 99.56 & 91.99 & \textbf{88.87} & 99.29 & \textbf{77.39}  \\
		
		\bottomrule
	\end{tabular}
\end{table}

\subsection{More Cross-manipulation Results}
To further demonstrate the generalization ability of our proposed method among different manipulated types, we show more cross-manipulation results on FF++. 
As shown in Table \ref{tab:supp-cross-man}, our method consistently surpasses all competitors by a clear margin in most cases.
These results demonstrates that our CDFA improving the cross-manipulation performance.

\subsection{More Visualization Results}
\begin{figure*}[t]
	\centering
	\subfloat[Xcep on CDF]{\includegraphics[width=0.22\linewidth]{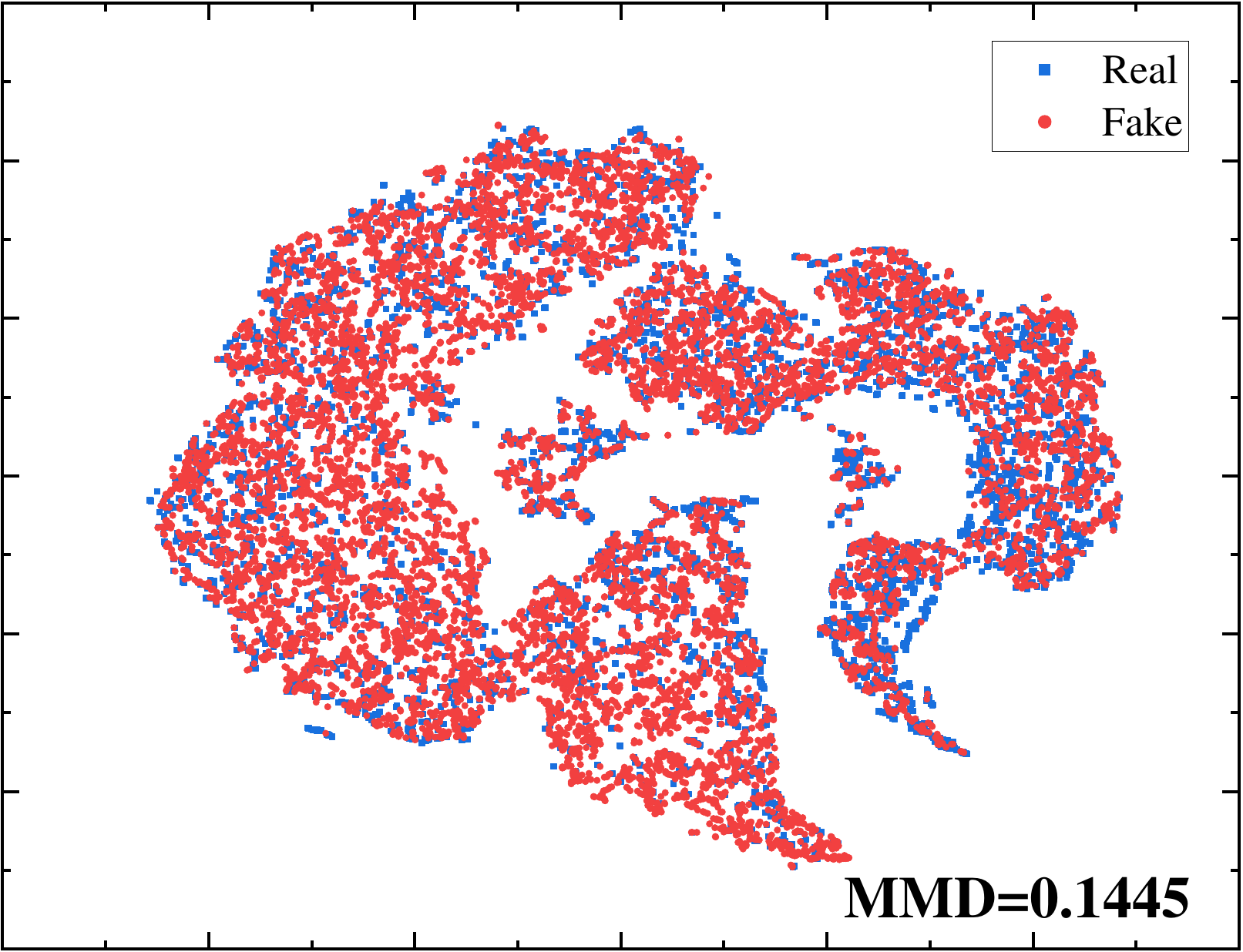} \label{xce_celeb}}
	\hfill
	\subfloat[Xcep+CDFA on CDF]{\includegraphics[width=0.22\linewidth]{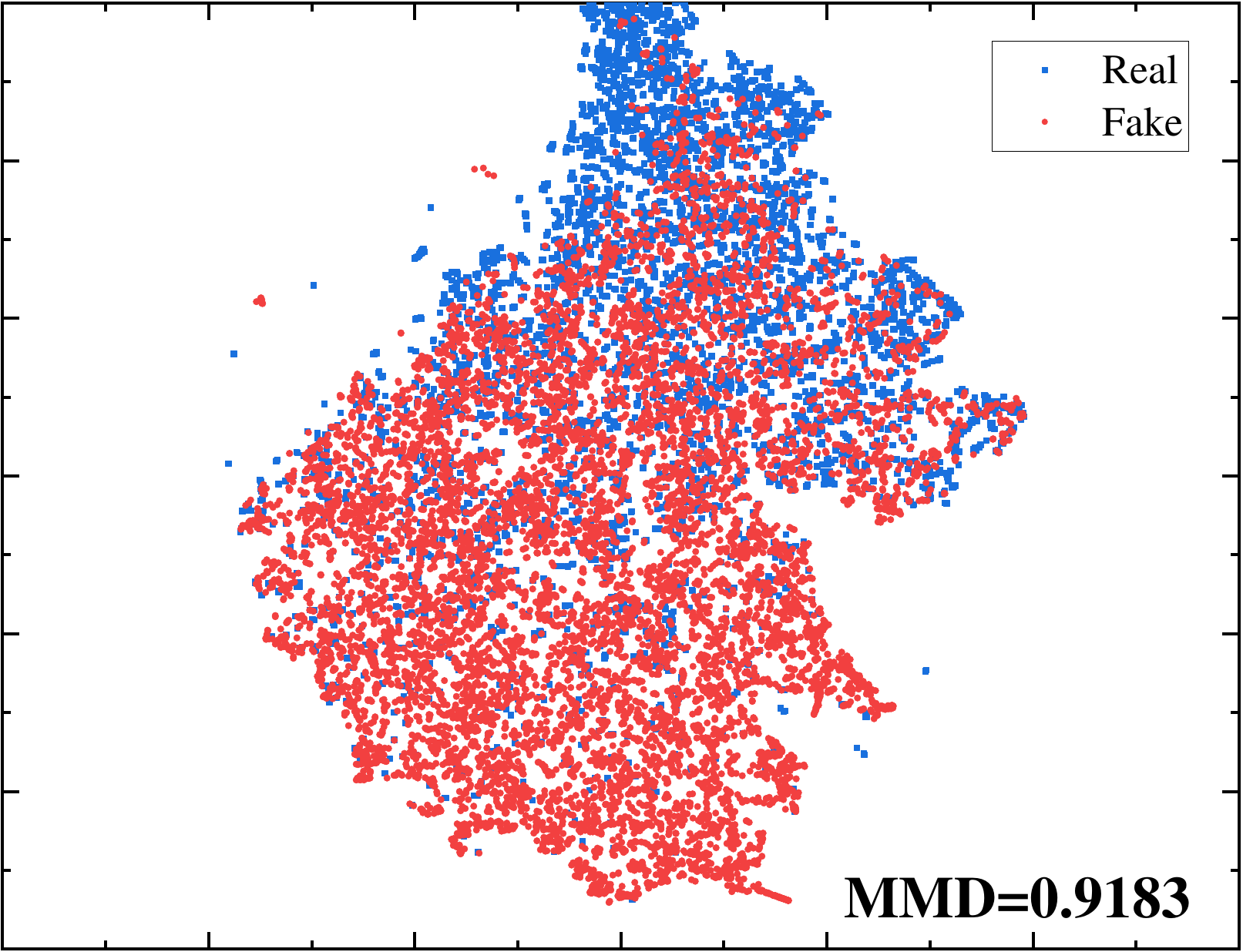} \label{xce_our_celeb}}
	\hfill
	\subfloat[Xcep on Wild]{\includegraphics[width=0.22\linewidth]{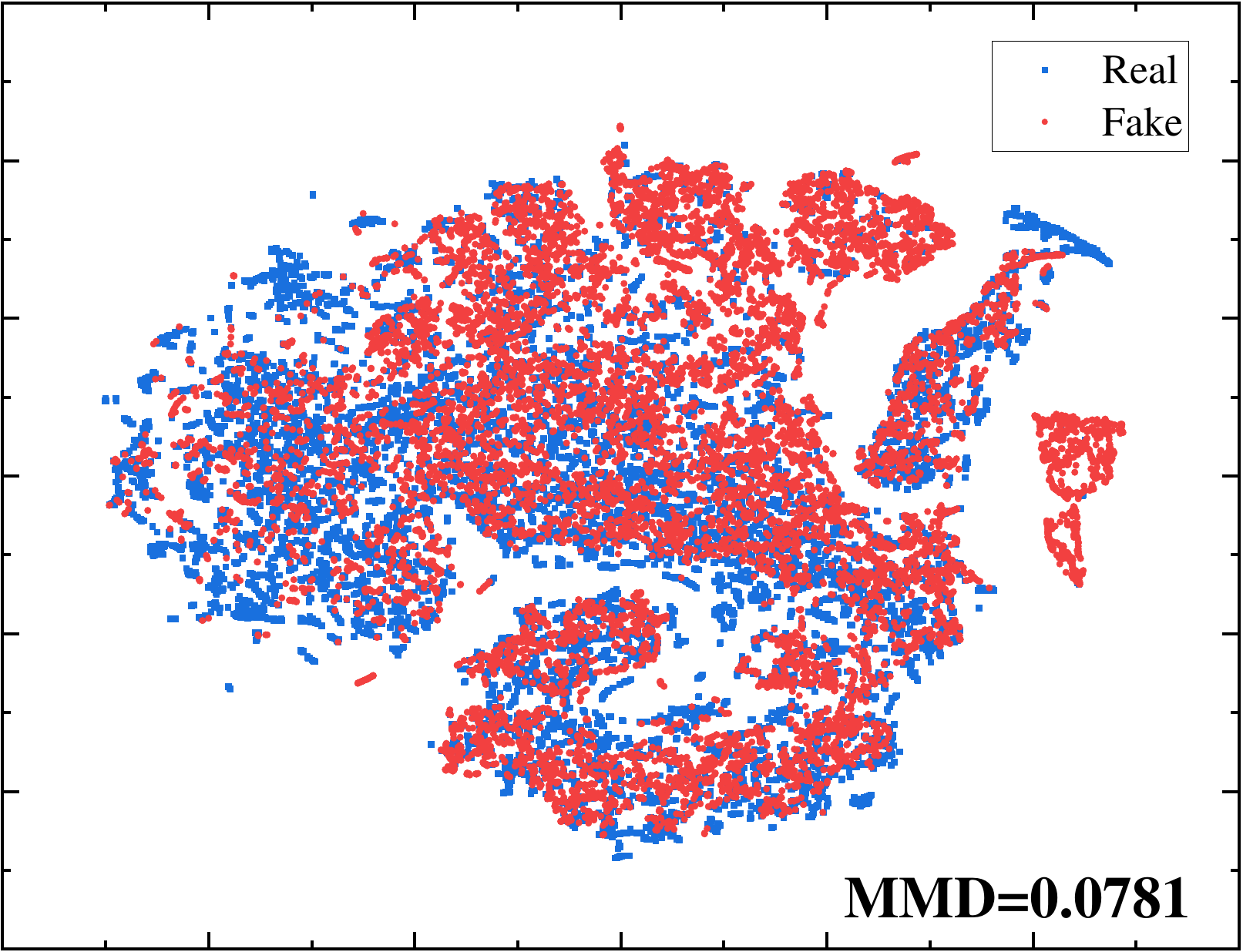} \label{xce_wild}}
	\hfill
	\subfloat[Xcep + CDFA on Wild]{\includegraphics[width=0.22\linewidth]{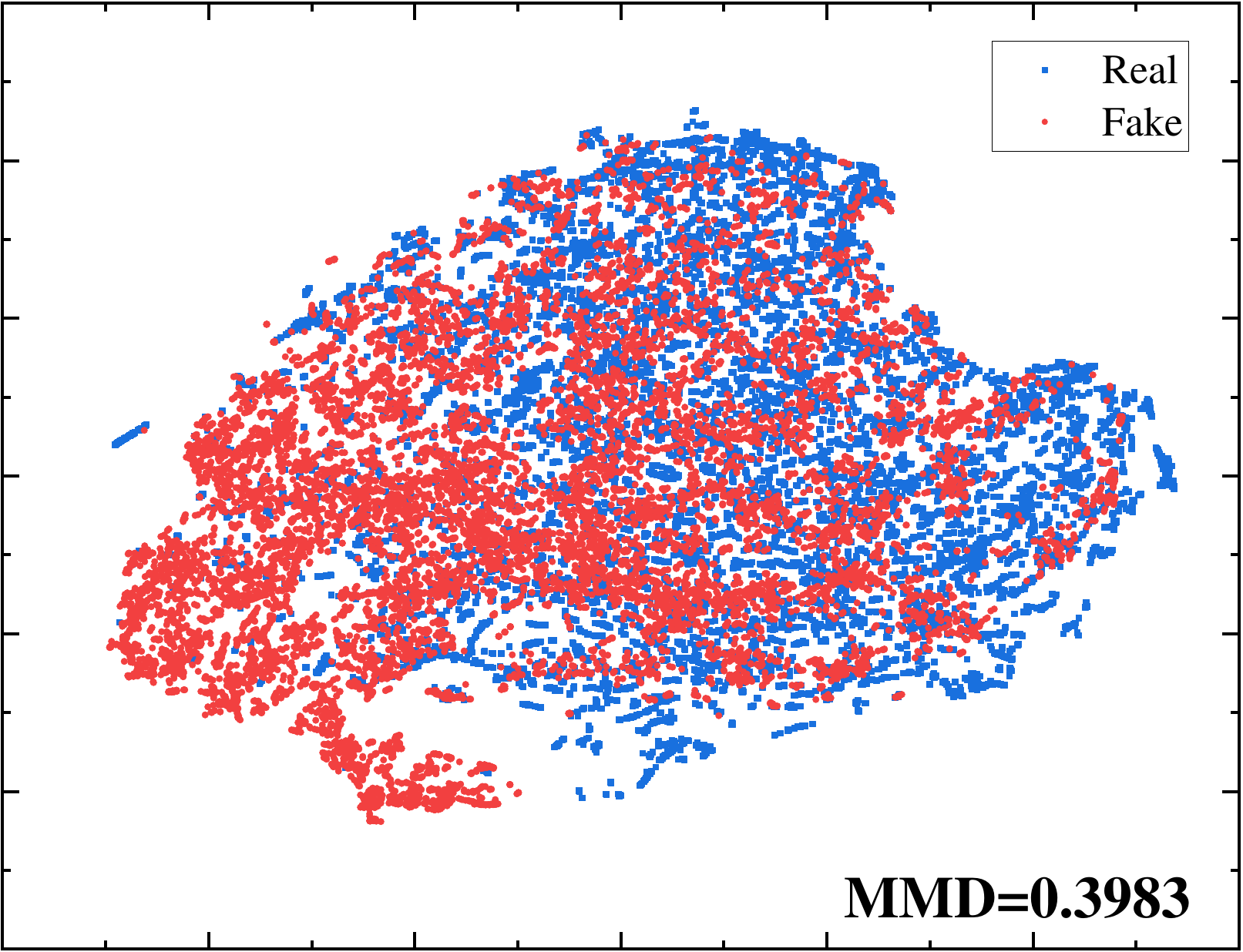} \label{xce_our_wild}}
	\hfill
	\subfloat[ENb4 on CDF]{\includegraphics[width=0.22\linewidth]{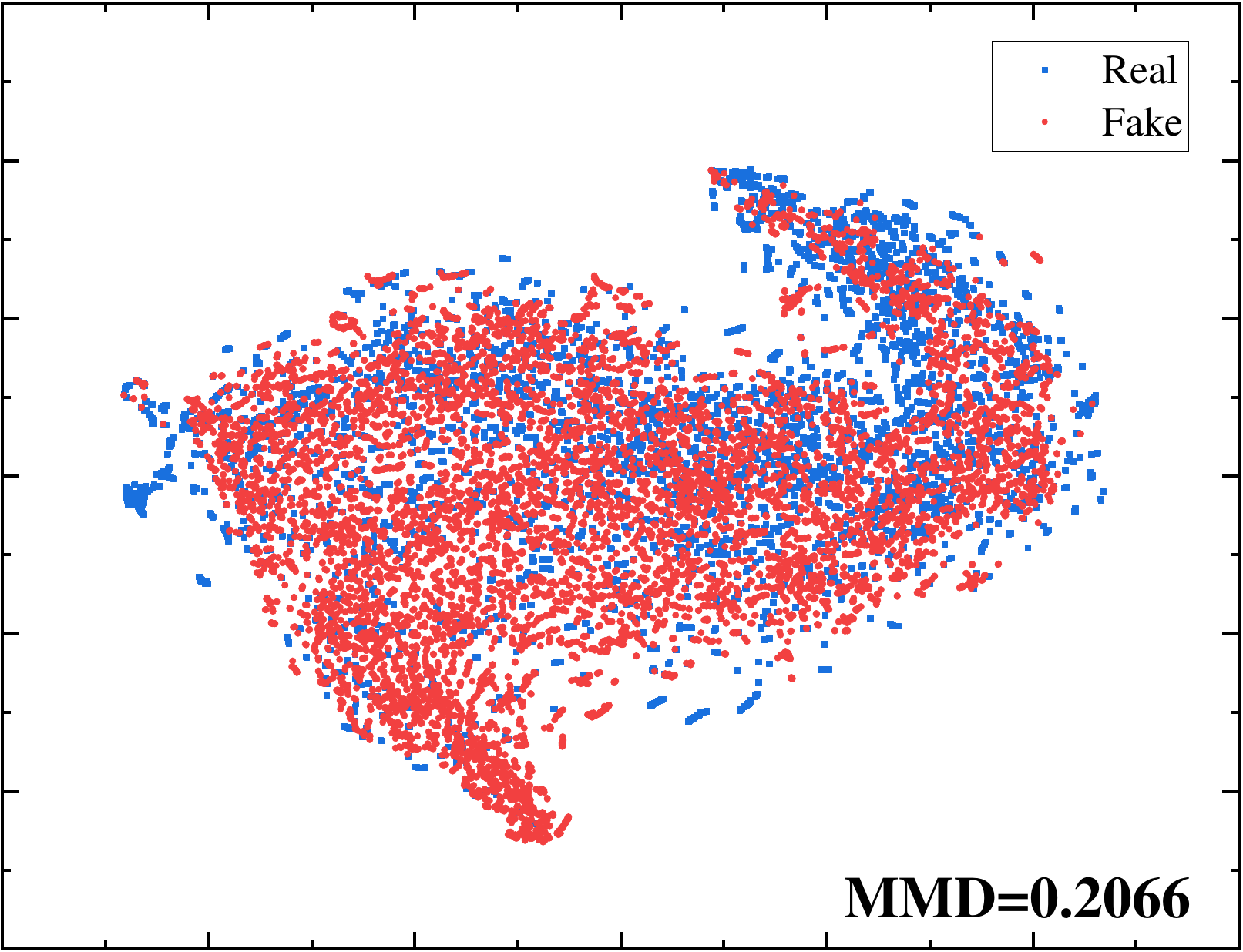} \label{efn_celeb}}
	\hfill
	\subfloat[ENb4 + CDFA on CDF]{\includegraphics[width=0.22\linewidth]{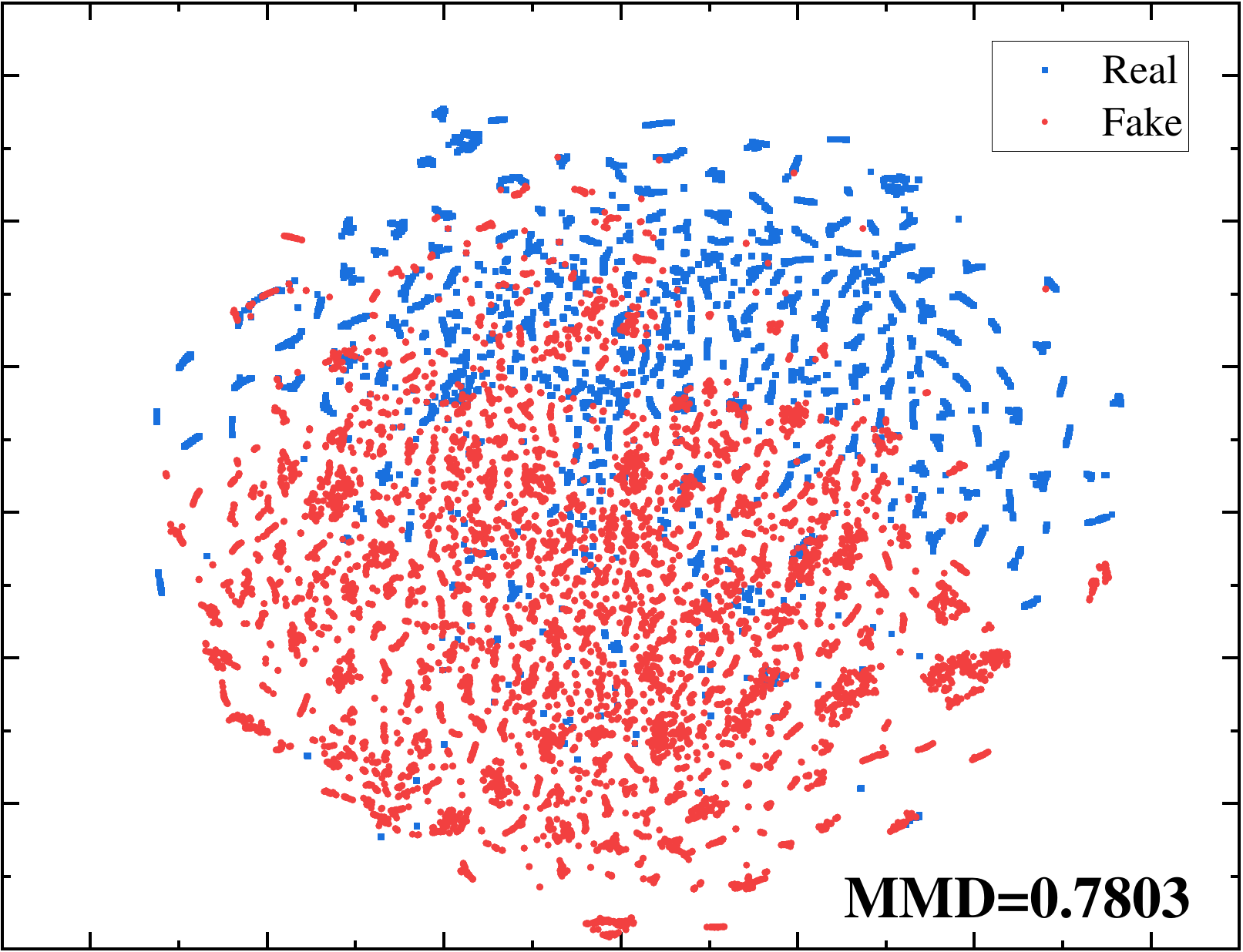} \label{efn_our_celeb}}
	\hfill
	\subfloat[ENb4 on Wild]{\includegraphics[width=0.22\linewidth]{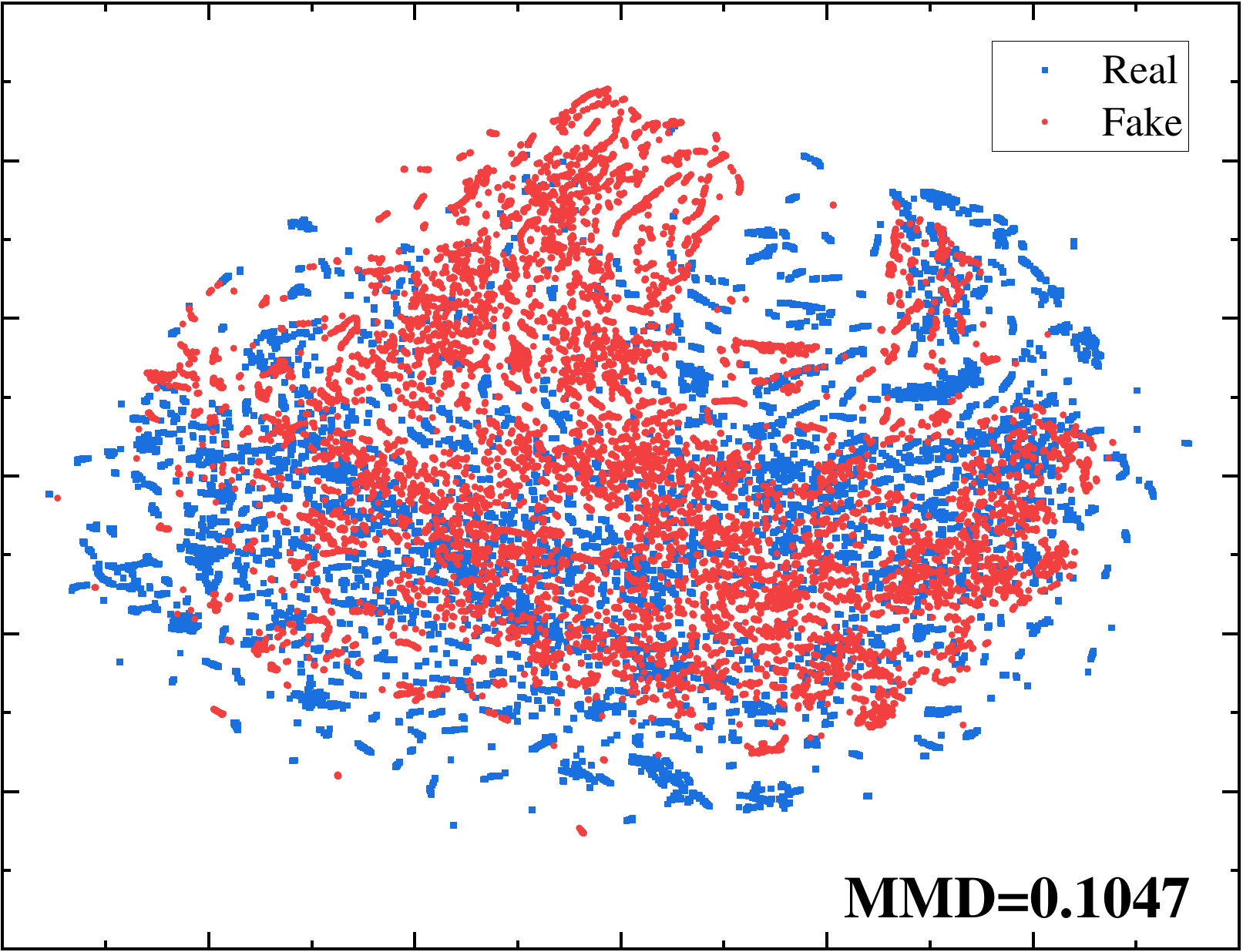} \label{efn_wild}}
	\hfill
	\subfloat[ENb4 + CDFA on Wild]{\includegraphics[width=0.22\linewidth]{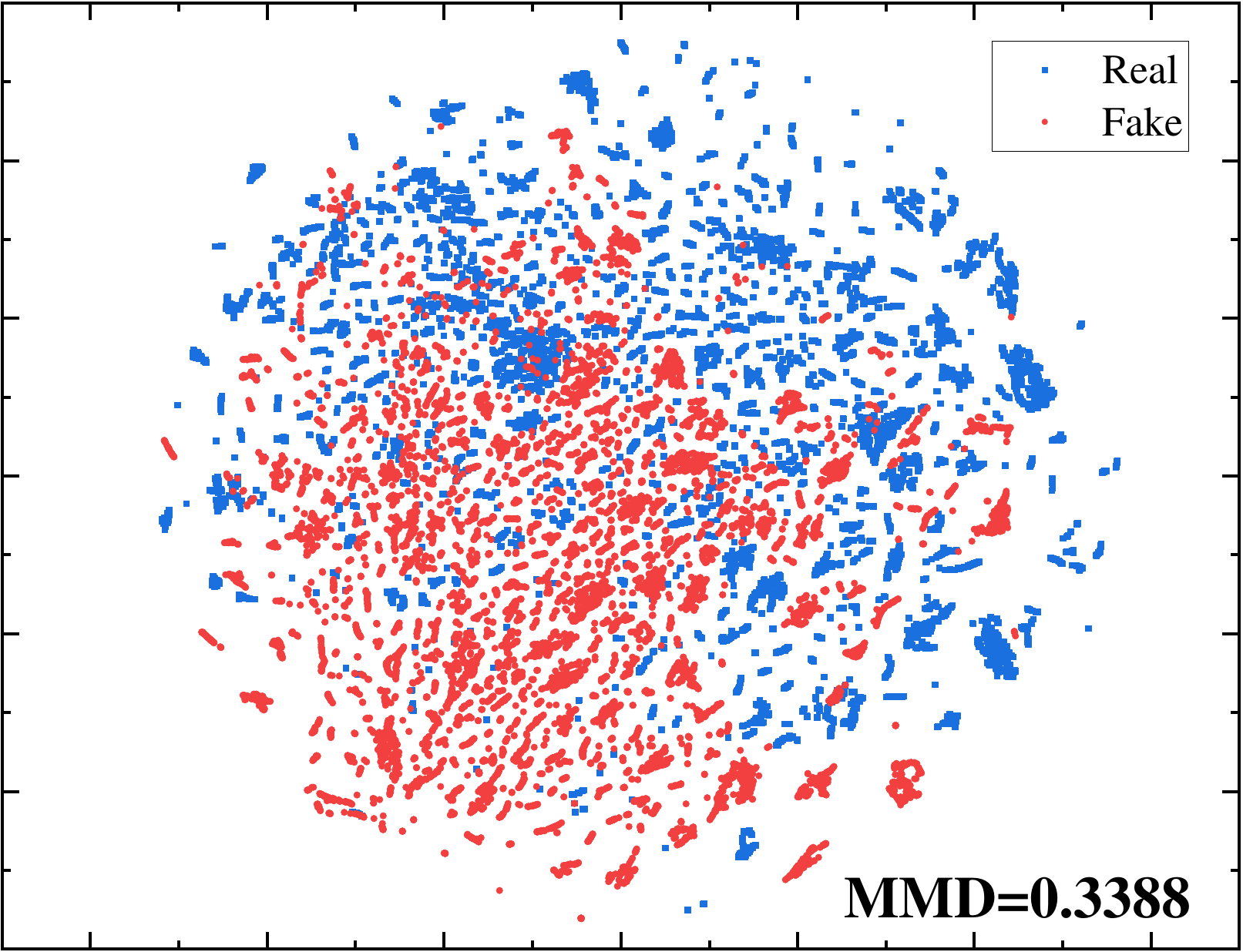} \label{efn_our_wild}}
	\hfill
	\subfloat[Swin on CDF]{\includegraphics[width=0.22\linewidth]{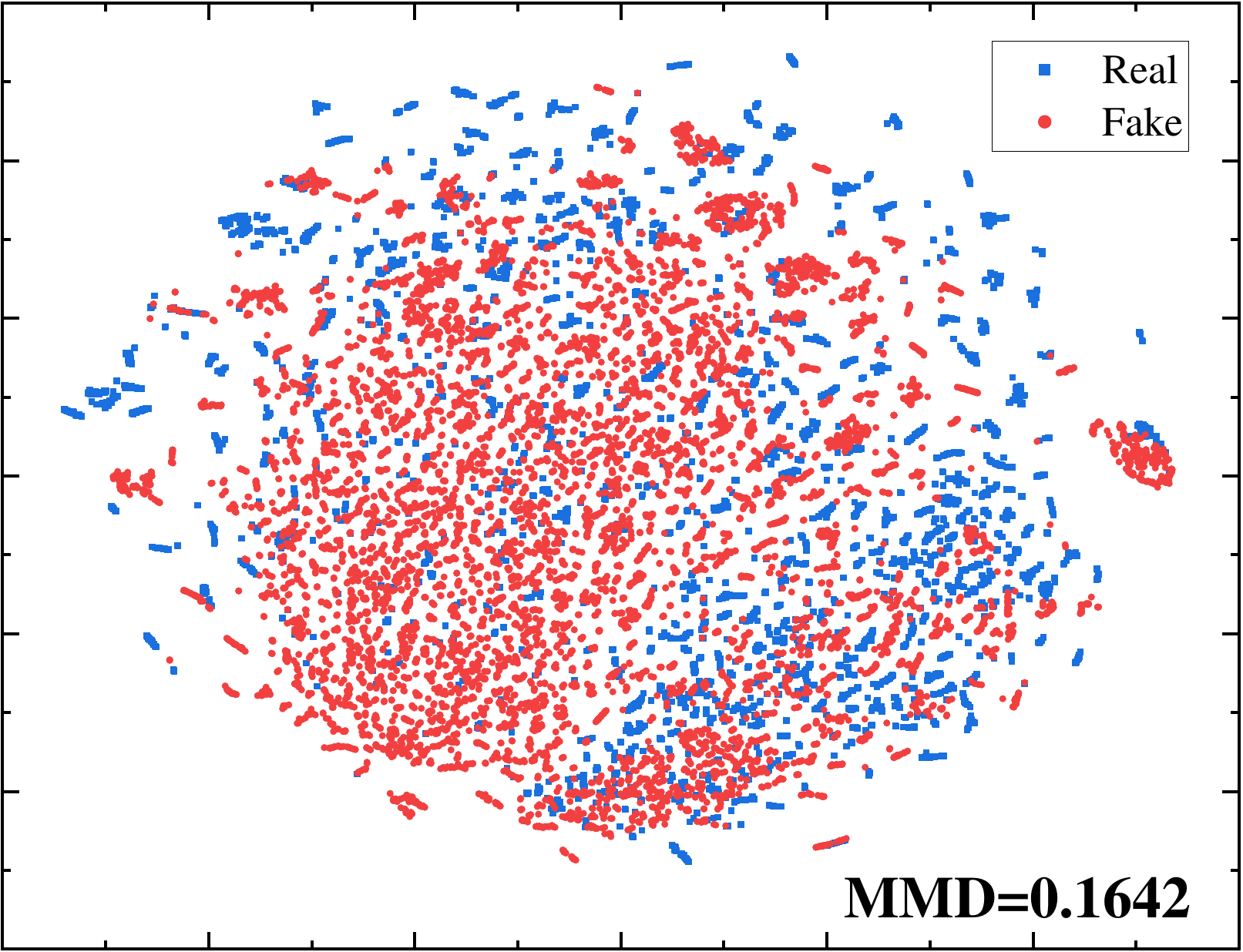} \label{swin_celeb}}
	\hfill
	\subfloat[Swin + CDFA on CDF]{\includegraphics[width=0.22\linewidth]{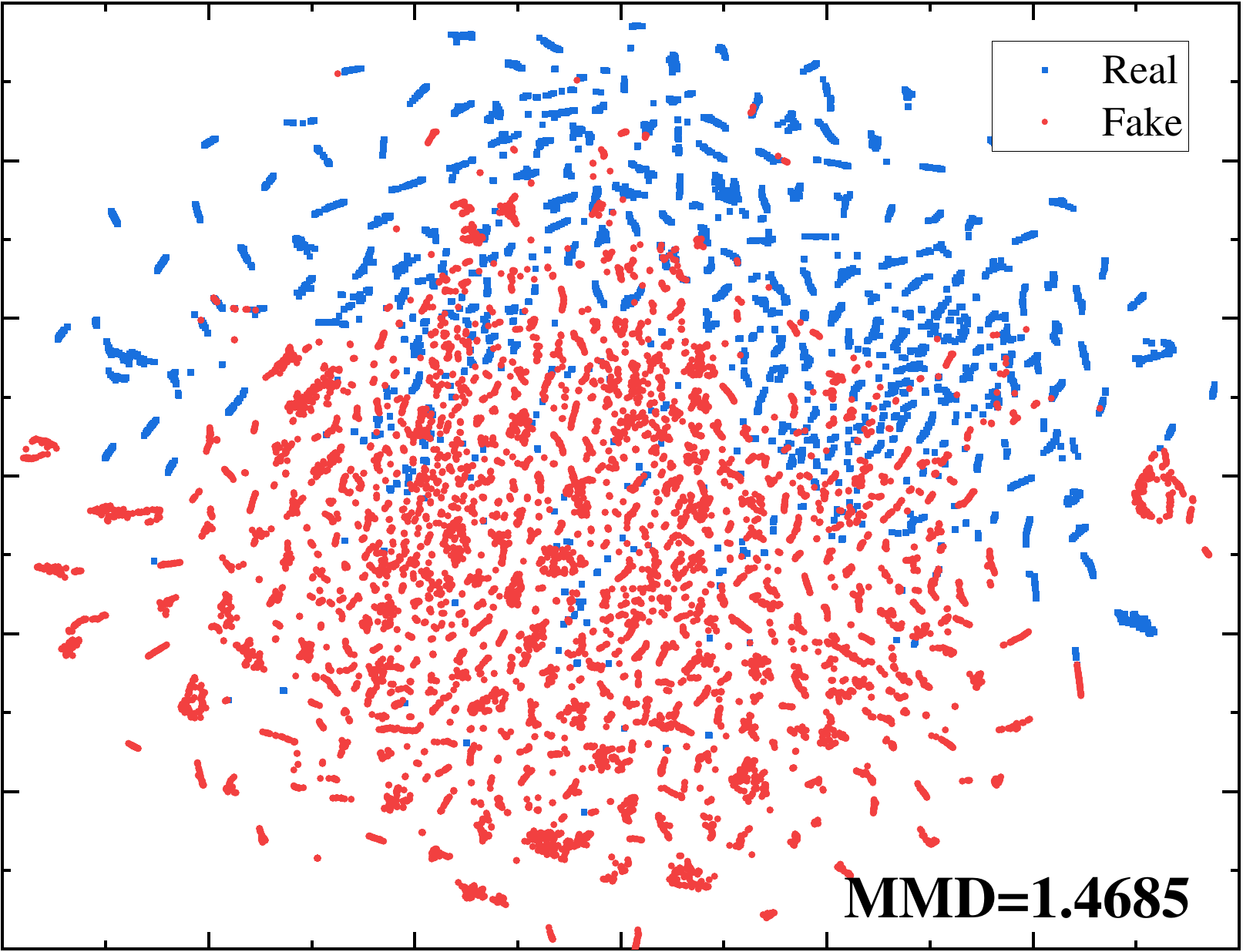} \label{swin_our_celeb}}
	\hfill
	\subfloat[Swin on Wild]{\includegraphics[width=0.22\linewidth]{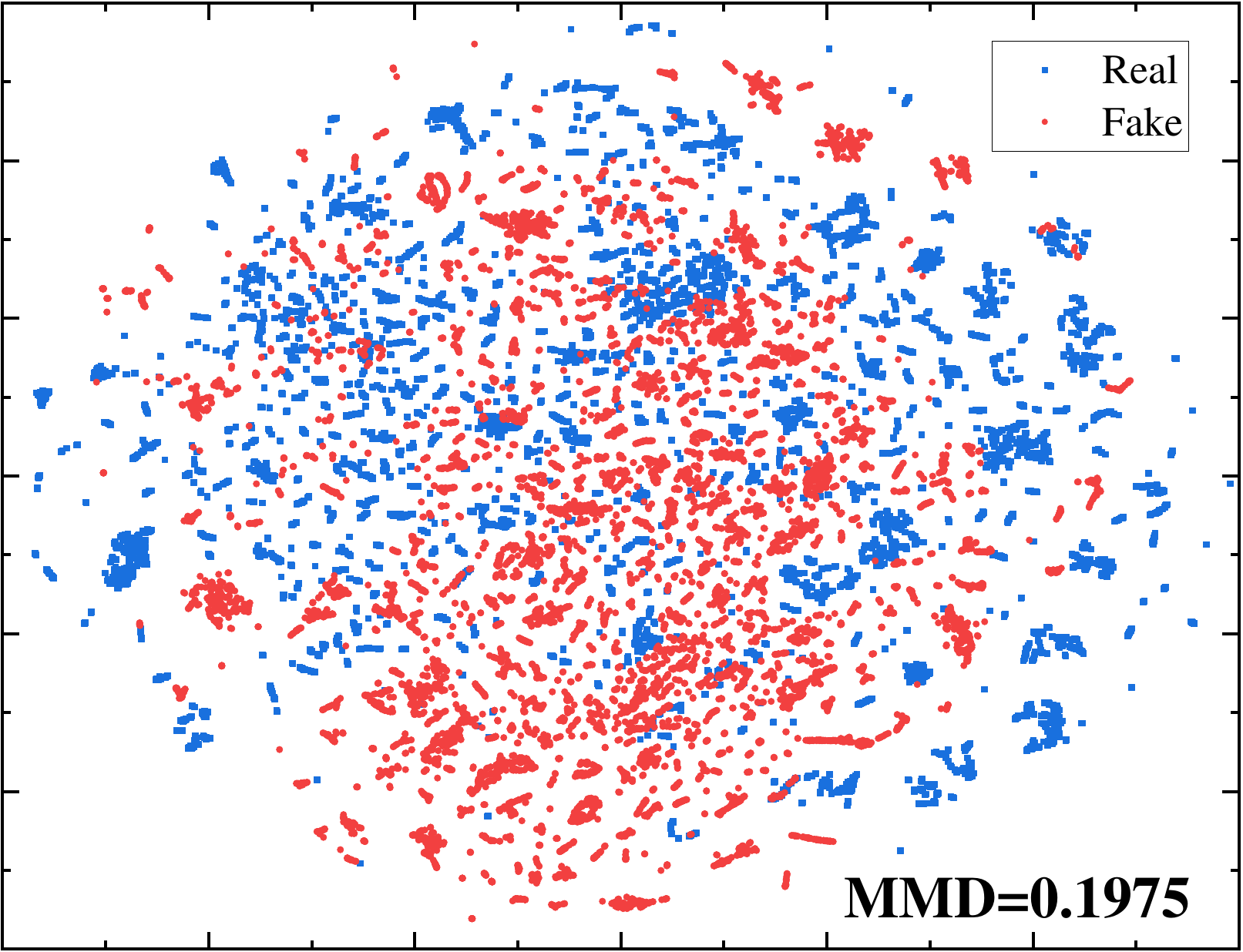} \label{swin_wild}}
	\hfill
	\subfloat[Swin + CDFA on Wild]{\includegraphics[width=0.22\linewidth]{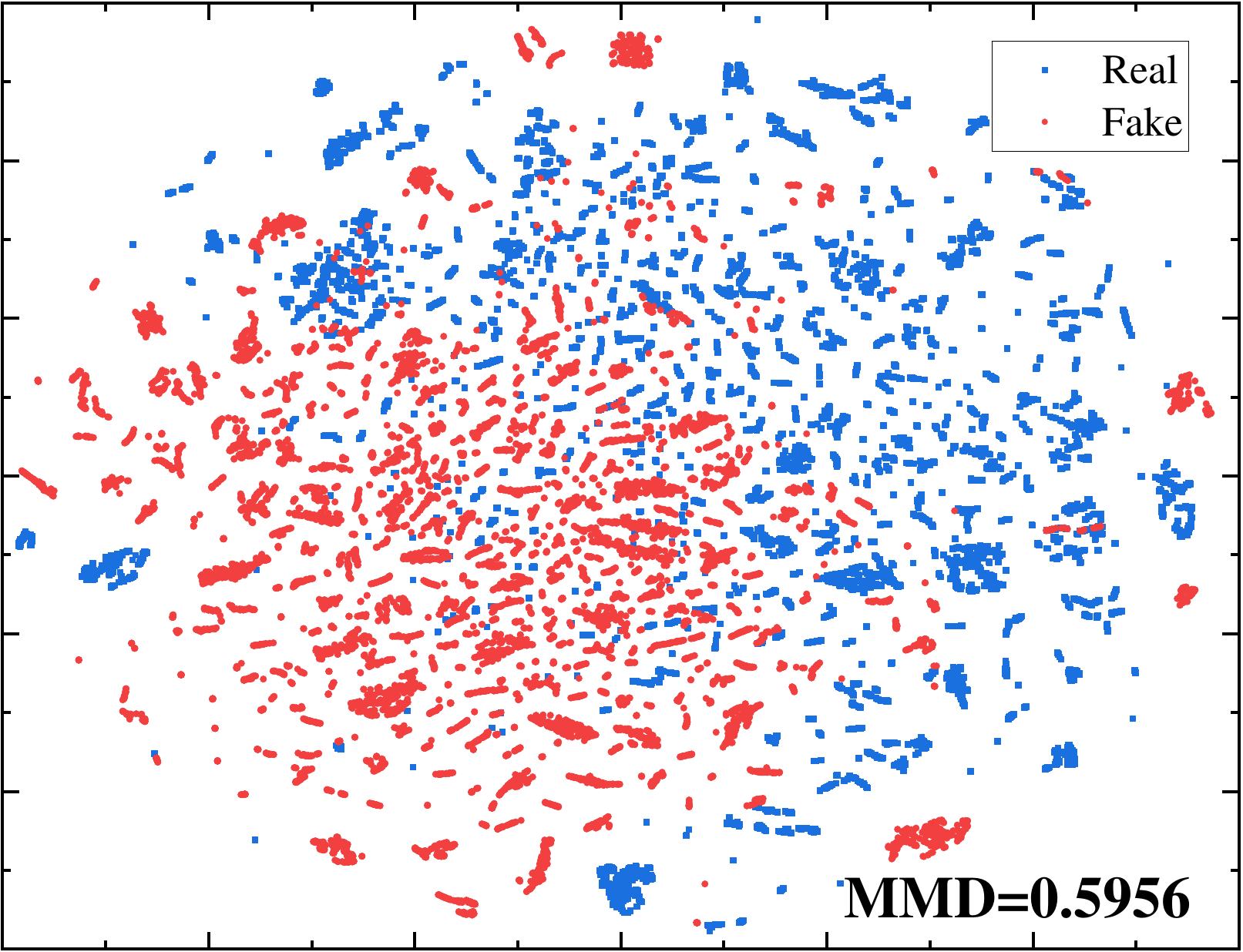} \label{swin_our_wild}}
	\caption{Feature visualization of the cross-datasets evaluation. Trained on FF++.}
	\label{fig:supp-tsne}
\end{figure*}

We apply the t-SNE method for visualizing features from the last layers of $f_{\alpha}$. Moreover, we compute the Maximum Mean Discrepancy (MMD) distance to evaluate the gap of feature distributions. A larger MMD indicates that the distributions of two data are more different.
As shown in Figure \ref{fig:supp-tsne}, we can observe that when testing unseen deepfakes, adding the proposed CDFA significantly enhances the distinction in feature distribution between real and fake faces extracted by the deepfake detector, thereby improving its generalization performances.

\subsection{More Evaluation about DFS}
To further validate the effectiveness of DFS, we conduct additional experiments that only train the models with real faces. 
In this scenario, the fake part of the training data is the p-fake samples, and thus MC is not available. 
We explore two policies for utilizing three forgery augmentation operations, 1) fixed with uniform probabilities and 2) optimized by DFS.

As shown in Table \ref{tab:supp-dfs}, we can observe that DFS performs better than fixed policy when training on the real part of FF++ (FF++-real) and evaluating the subset of five manipulations. It demonstrates that the forgery artifacts simulated by DFS are more diverse than that by the fixed policy.
When we changed the training data to the real part of CelebDF (CDF-real), the performances of fixed policy and DFS on FF++ subsets suffered from a significant drop due to the mismatch of data sources. But DFS still performs better than fixed policy. It further proves that DFS can simulate more general forgery artifacts by optimizing the augmentation policy during the training.

\begin{table}[h]
	\centering
	\caption{Video-level AUC(\%) on cross-manipulation evaluations under the real training scenario. The best results are highlighted.}
	\label{tab:supp-dfs}
	\begin{tabular}{cccccccl} 
		\toprule
		Policy & Training Data & DF    & F2F   & FS    & NT    & FSh   & Avg    \\ 
		\midrule
		DFS    & FF++-real    & 99.41 & \textbf{95.47} & \textbf{95.86} & \textbf{91.45} & \textbf{85.63} & \textbf{93.56}  \\
		Fixed & FF++-real    & \textbf{99.56} & 89.18 & 91.54 & 85.82 & 76.68 & 88.56  \\
		\midrule
		DFS    & CDF-real       & 92.94 & \textbf{74.60} &\textbf{ 88.99} & \textbf{74.77} & \textbf{64.90} & \textbf{79.24}  \\
		Fixed  & CDF-real       &\textbf{93.86}       &70.84       &76.55       &71.53       &55.84       &73.72       \\
		\bottomrule
	\end{tabular}
\end{table}

\section{Limitations} 
Although our results in cross-dataset and cross-manipulation evaluations are expected to be beneficial, we observe some limitations of our method. Similar to other forgery augmentation methods \cite{FaceXRayMore2020li,DetectingDeepfakesSelfBlended2022shiohara}, our method does not perform well on whole-image synthesis because we define a “fake image” as an image where the face region is manipulated. We believe that our CDFA is expected to further benefit from future developments in forgery augmentation topologies.
	
\end{document}